\documentclass[10pt,twocolumn,letterpaper]{article}

\usepackage{cvpr}

\definecolor{cvprblue}{rgb}{0.21,0.49,0.74}
\usepackage[pagebackref,breaklinks,colorlinks,allcolors=cvprblue]{hyperref}

\def\paperID{} % 提交系统分配的 ID
\usepackage[utf8]{inputenc}
\usepackage[T1]{fontenc}
\usepackage{booktabs}      % professional tables
\usepackage{amsfonts}      % blackboard math symbols
\usepackage{amsmath}       % math
\usepackage{amssymb}       % more math symbols
\usepackage{mathtools}     % extensions to amsmath
\usepackage{nicefrac}      % compact fractions
\usepackage{microtype}     % typographic refinements
\usepackage[dvipsnames]{xcolor}
\usepackage[table]{xcolor} % 导言区
\usepackage{enumitem}

\usepackage{graphicx}
\usepackage{algorithm}
\usepackage{algorithmic}
\usepackage{multirow}
\usepackage{array}
\usepackage{caption}
\usepackage{titletoc}
\usepackage{amsthm}

\usepackage{subcaption}

\usepackage{amsmath,amsfonts,bm}

\def\eqref#1{equation~\ref{#1}}
\def\1{\bm{1}}

\DeclareMathAlphabet{\mathsfit}{\encodingdefault}{\sfdefault}{m}{sl}
\SetMathAlphabet{\mathsfit}{bold}{\encodingdefault}{\sfdefault}{bx}{n}

\definecolor{level4}{HTML}{283593}      % 深蓝紫 >=80%
\definecolor{level3}{HTML}{009688}       % 青绿 70–80%
\definecolor{level2}{HTML}{00C853}   % 鲜亮的正绿 60–70%
\definecolor{level1}{HTML}{69FB6A}    % 荧光浅绿 50–60%

\theoremstyle{plain}
\newtheorem{theorem}{Theorem}[section]
\newtheorem{lemma}[theorem]{Lemma}

\theoremstyle{definition}

\newtheorem{assumption}[theorem]{Assumption}

\theoremstyle{remark}
\newtheorem{remark}[theorem]{Remark}

\title{IPCV: Information-Preserving Compression for MLLM Visual Encoders}

\author{
Yuan Chen$^{1,2}$\thanks{Equal Contribution.} \quad
Zichen Wen$^{1,3*}$ \quad
Yuzhou Wu$^{1,4}$ \quad
Xuyang Liu$^{1,5}$ \quad
Shuang Chen$^{1}$ \\
Junpeng Ma$^{6}$ \quad
Weijia Li$^{7,3}$\quad
Conghui He$^{3}$\quad
Linfeng Zhang$^{1}$\thanks{Corresponding author: \texttt{zhanglinfeng@sjtu.edu.cn}} \\
$^{1}$ EPIC Lab, SJTU \quad
$^{2}$ CityU \quad
$^{3}$ Shanghai AI Laboratory \\
$^{4}$ University of Sheffield \quad
$^{5}$ SCU \quad
$^{6}$ FDU \quad
$^{7}$ SYSU
}

\date{}  % 采用提交时的系统日期，留空即可

\begin{document}

\maketitle

\begin{abstract}
% To overcome these challenges, we introduce IPCV, a \textbf{training-free, information-preserving compression} framework for MLLM visual encoders. IPCV facilitates aggressive token pruning within the ViT by employing a novel \textbf{Neighbor-Guided Reconstruction (NGR)} mechanism. NGR allows pruned tokens to be temporarily reconstructed to participate in attention with minimal overhead, and then fully restores them before they enter the LLM. This semantic recovery step ensures that a complete and high-fidelity set of visual features is available for downstream multimodal reasoning. Paired with a lightweight LLM-stage pruner, IPCV's dual-stage design maximizes computational savings while preserving semantic integrity. Extensive experiments demonstrate that IPCV substantially reduces end-to-end computation and outperforms state-of-the-art training-free token compression methods on a wide range of image and video benchmarks.
%
% Multimodal Large Language Models (MLLMs) deliver strong vision–language performance but incur high computational cost, driven by the large number of visual tokens processed by the Vision Transformer (ViT) encoder.
% Multimodal Large Language Models (MLLMs) deliver strong vision–language performance but at high computational cost, driven by the large number of visual tokens processed by the Vision Transformer (ViT) encoder.
% Multimodal Large Language Models (MLLMs) deliver strong vision-language performance but at high computational cost, driven by the many visual tokens processed by the Vision Transformer (ViT) encoder. 
Multimodal Large Language Models (MLLMs) deliver strong vision-language performance but at high computational cost, driven by numerous visual tokens processed by the Vision Transformer (ViT) encoder. 
Existing token pruning strategies are inadequate: LLM-stage token pruning overlooks the ViT's overhead, while conventional ViT token pruning, without language guidance, risks discarding textually critical visual cues and introduces feature distortions amplified by the ViT's bidirectional attention.
To meet these challenges, we propose \textbf{IPCV}, a training-free, information-preserving compression framework for MLLM visual encoders. IPCV enables aggressive token pruning inside the ViT via \textbf{Neighbor-Guided Reconstruction (NGR)} that temporarily reconstructs pruned tokens to participate in attention with minimal overhead, then fully restores them before passing to the LLM. 
Besides, we introduce \textbf{Attention Stabilization (AS)} to further alleviate the negative influence from token pruning by approximating the K/V of pruned tokens. It can be directly applied to previous LLM-side token pruning methods to enhance their performance.
Extensive experiments show that IPCV substantially reduces end-to-end computation and outperforms state-of-the-art training-free token compression methods across diverse image and video benchmarks. Our code is available at \url{https://github.com/Perkzi/IPCV}.
%Our code will be released on GitHub.
% Our code is provided in the supplementary material.

\end{abstract}

% 1. 现在的mllm token prune基本都只能做llm阶段，做不了vit阶段，以前vit的方法在mllm上效果也不好。但是实际计算中mllm的vit部分计算成本是很高的。
% 2. mllm中vit阶段token压缩的问题在于，这一阶段没有language信息参与，很有可能会把vision中不重要，但是language上重要的信息给删除掉。同时，被删除的vision token也会导致没有被删除的vision token的特征存在误差。这一点在vit阶段比llm阶段更严重，因为vit是双向注意力，影响范围更大。
% 3.我们的方法: 删除vit token，但是在llm前恢复，防止language相关信息被删除。
\section{Introduction}
\begin{figure}[htbp]
    \centering
    \includegraphics[width=0.48\textwidth]{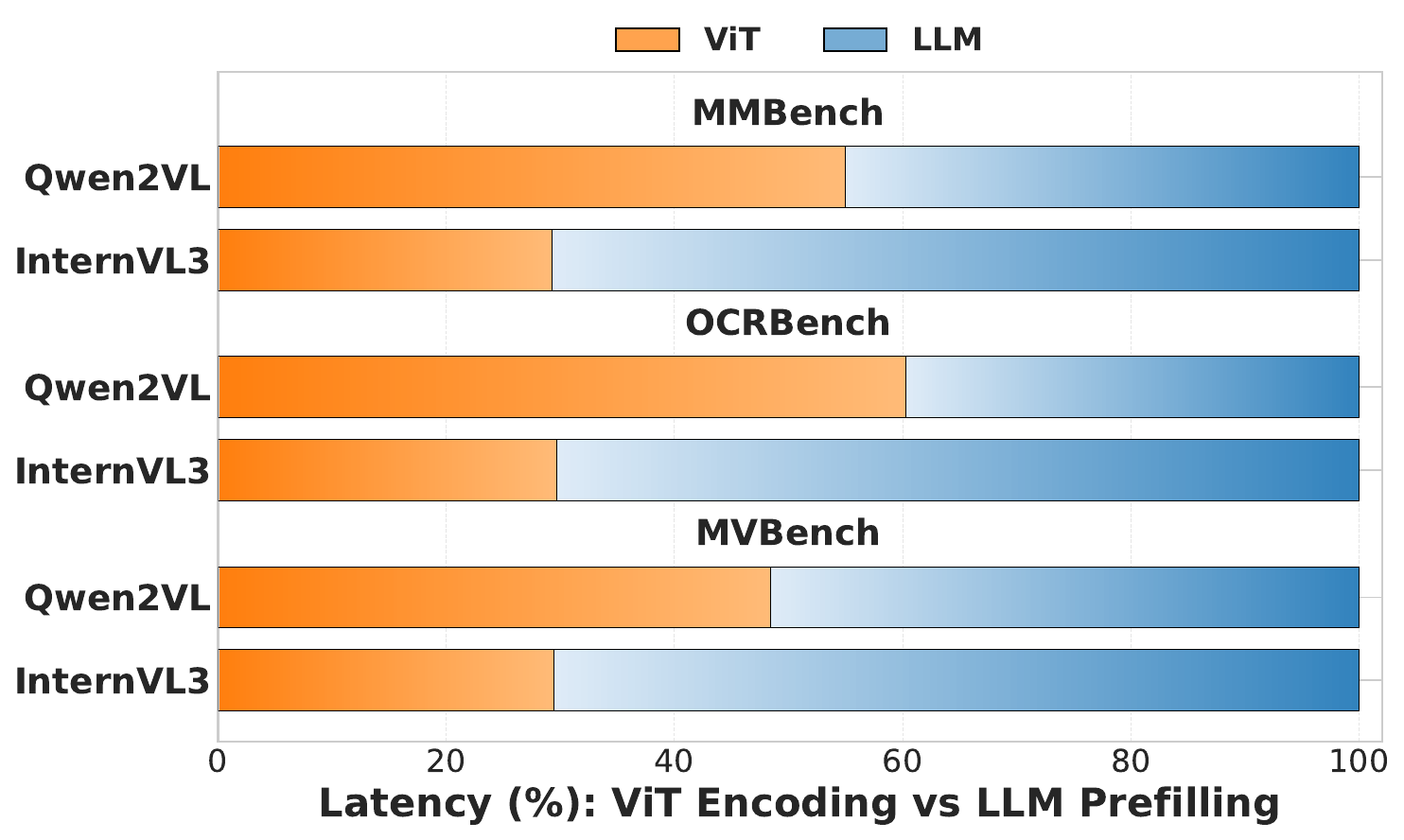}
    % \vspace{-0.2em}
    \caption{\textbf{ViT Encoding versus LLM Prefilling latency proportions} across three benchmarks---MMBench (low-resolution images), OCRBench (high-resolution images), and MVBench (videos)---using Qwen2-VL-7B-Instruct and InternVL3-38B.}
    % \vspace{-0.4cm}
    \label{fig:latency_proportion}
    \vspace{-1.0em}
  \end{figure}
Multimodal large language models (MLLMs)~\cite{liu2023llava,qwen2vl} have achieved remarkable performance across diverse vision-language tasks~\cite{wang2024internvideo2,wang2024exploring,wen2025ai,zhang2025docr,zhang2025trivia,chen2024mj,Zhang_2025_ICCV,kang2025legion}, benefiting from powerful language reasoning capabilities and strong vision encoders~\cite{radford2021learning, zhai2023sigmoid}. 
However, the visual branch of these models often produces hundreds to thousands of tokens per image or video, especially at high resolutions or with multiple frames~\cite{cui2025vico}. 
Due to the quadratic computational complexity $\mathcal{O}(N^2)$ of self-attention~\cite{vaswani2017attention}, longer token sequences incur significantly higher latency. This severely hinders the widespread adoption and edge deployment of MLLMs.
In practice, however, visual tokens in MLLMs exhibit substantial redundancy~\cite{marr2010vision}. To address this, a surge of recent research has focused on token pruning methods~\cite{liu2025shifting,liu2024multi,wen2025efficient,liu2025mixing,liu2025global}, which aim to directly reduce the number of tokens and lower computational costs.

Nevertheless, despite the remarkable progress of these approaches, most operate only on the LLM within MLLMs. In practice, as vision encoders increasingly adopt dynamic designs or higher native resolutions, the resulting visual token sequences grow significantly longer, rendering the computational cost of the vision encoder impossible to overlook (as shown in Figure~\ref{fig:latency_proportion}). 
While several methods have been proposed for token compression at the vision transformer (e.g., ToMe~\cite{bolya2023tome}, ToFu~\cite{kim2024token}), they primarily focus on vision-only models and tasks (e.g., classification), resulting in suboptimal performance on multimodal understanding tasks~\cite{zhang2024sparsevlm}. Generally speaking, token pruning in the ViT for MLLMs faces two significant challenges.

\begin{figure*}[t]
    \centering
    \includegraphics[width=1\linewidth]{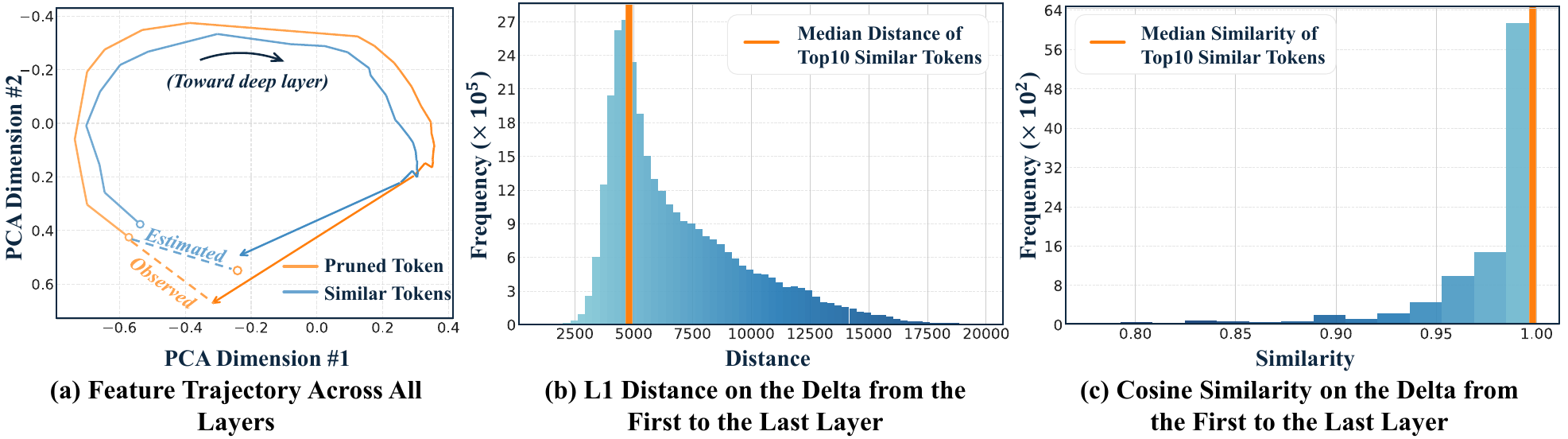}
    % \vspace{-1.8em}
    \caption{\textbf{Visualization of the delta of tokens from the shallow layers to the deep layers.}
    (a) PCA projection of the hidden state trajectories of a pruned token and the mean of its top-10 nearest neighbors, traced from the pruning layer to the final layer of the ViT, with arrows indicating the shift direction. 
    (b) and (c): Distributions of L1 distance and cosine similarity, computed from pairwise comparisons between the change (delta) of retained tokens and that of pruned tokens across layers. 
    The deep-blue vertical line denotes the overall median of the pairwise distances (cosine similarities) computed between each pruned token and the mean of its top-10 most similar tokens.
These visualizations reveal that tokens with high similarity tend to exhibit highly similar changes from shallow to deep layers.
}
\label{fig:trajectory_frequency}
\vspace{-1.4em}
\end{figure*}

\noindent\textbf{Absence of language guidance.}
A fundamental challenge of performing token compression at the vision encoder in MLLMs lies in the absence of language guidance~\cite{wen2025token}. Since the token compression process occurs before any cross-modal interaction, it risks discarding visual tokens that may appear unimportant from a purely visual perspective but are in fact crucial for downstream language-driven reasoning. 

\noindent\textbf{Negative influence from pruned tokens. } Moreover, the removal of certain vision tokens can introduce feature distortions in the remaining tokens, as the bidirectional attention mechanism in vision transformers~\cite{dosovitskiy2021image} propagates information globally. This issue is even more pronounced at the vision encoding stage than in the language model, where unidirectional attention limits the scope of such errors~\cite{nam2016dual,shao2024localglobal}. As a result, naive token pruning in the vision tower can inadvertently undermine the model's ability to preserve information essential for multimodal understanding. 

To tackle the above two challenges, we propose \textbf{Information-Preserving Compression} for MLLM Visual Encoders (\textbf{IPCV}).
Specifically, in the vision encoder (i.e., ViT), IPCV prunes tokens in shallow layers to reduce the computation costs. 
Then, in the final layer, we introduce Neighbor-Guided Reconstruction (NGR), which aims to reconstruct the pruned tokens in the following layers based on their most similar remaining tokens (i.e., tokens that are not pruned). 
As studied in Figure~\ref{fig:trajectory_frequency}, we find that similar tokens exhibit similar delta from the shallow layers to the deep layers. As a result, instead of directly copying the similar remaining tokens~\cite{bolya2023tome}, NGR reconstructs the pruned tokens in the final layer by adding their values in the shallow layers with the delta of their most similar and remaining tokens from the shallow layers to the deep layers.

Secondly, to further reduce the negative influence of the pruned tokens on the remaining tokens, we introduce Attention Stabilization (AS), which aims to approximate the keys and values of pruned tokens in the middle layers, which is achieved by reconstructing pruned tokens via NGR, and then computing their keys and values. 

The proposed IPCV is training-free for the ViT and integrates seamlessly into diverse MLLMs. For the LLM stage, IPCV is orthogonal to and compatible with various token compression schemes; in this paper, we pair IPCV by default with the LLM-stage compressor in DART~\cite{wen2025dart} to further reduce redundancy after modal fusion.
In summary, our main contributions are as follows:
\begin{itemize}
[leftmargin=10pt, topsep=0pt, itemsep=1pt, partopsep=1pt, parsep=1pt]
    \item We provide a systematic comparison of LLM-stage and ViT-stage token pruning in MLLMs, revealing the need and challenges for vision-side token pruning in MLLMs.
    \item We propose \textbf{IPCV}, a training-free vision-side token pruning framework that uses Neighbor-Guided Reconstruction (NGR) to deliver a semantically complete vision token set to the LLM, and Attention Stabilization (AS) to mitigate negative influence from pruned tokens.
    \item Extensive experiments on various benchmarks with two representative MLLM models demonstrate our effectiveness across diverse tasks and settings.
\end{itemize}

\begin{figure*}[!t]
    \centering
    \includegraphics[width=\linewidth]{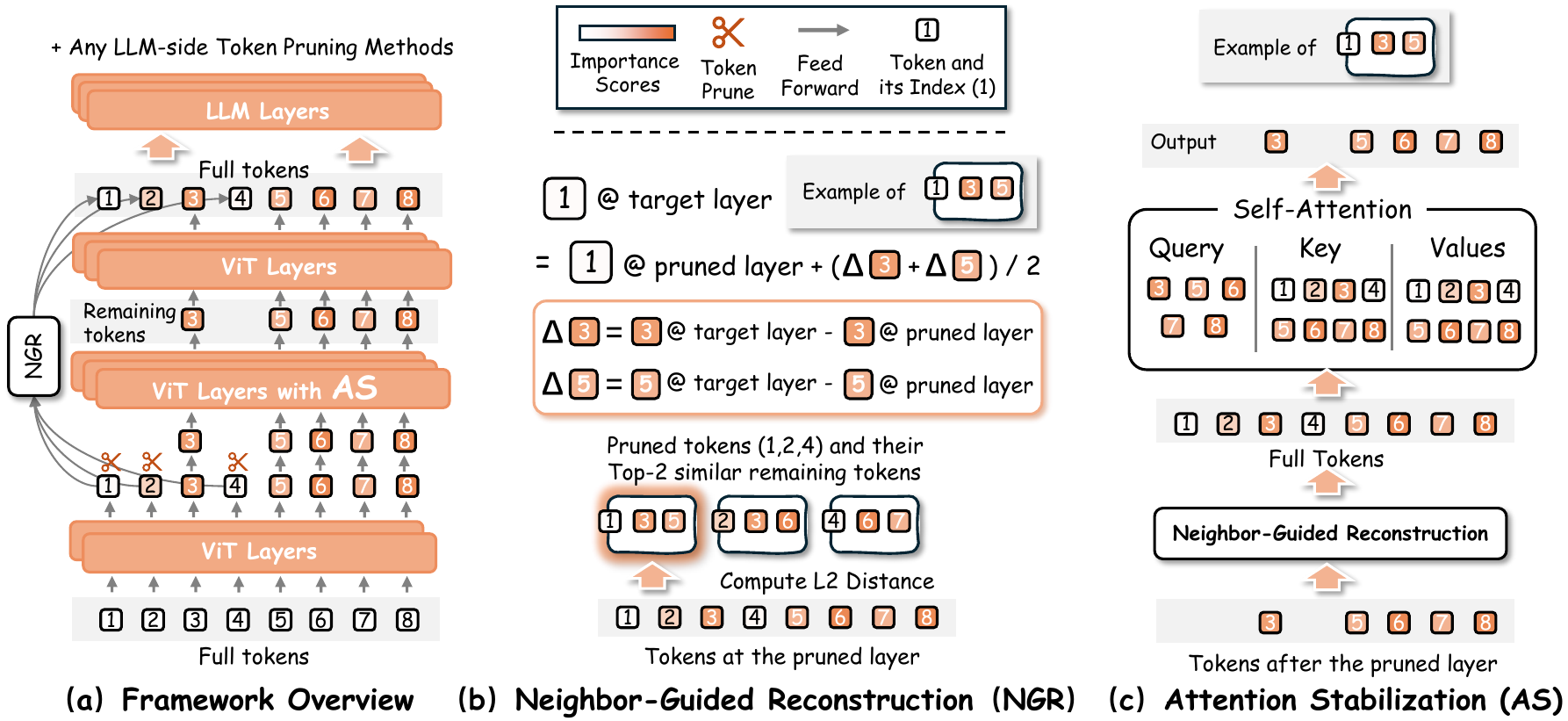} % 不用写后缀
    % \vspace{-1.8em}
    \caption{\textbf{Overview of IPCV.} IPCV prunes redundant visual tokens in the shallow layers of the vision encoder to reduce computation, then reconstructs the pruned tokens at the final layer using Neighbor-Guided Reconstruction (NGR) to deliver a semantically complete token set to the LLM. Attention Stabilization (AS) further mitigates the negative impact of token pruning by approximating the keys and values of the removed tokens in intermediate layers.}
    \label{fig:mylabel}
    \vspace{-1.0em}
\end{figure*}
\section{Related Work}

\noindent \textbf{Visual Compression in MLLM.}
% \paragraph{Visual Compression in MLLM.}
% To stem the inflow of visual tokens into the LLM, recent work compresses at multiple junctures.
% Recent work applies token compression at multiple points to reduce the number of visual tokens before they reach the LLM.
To reduce the inflow of visual tokens into the LLM, recent work compresses at multiple junctures~\cite{liao2025we,he2025audiomarathon,han2024ficoco,ma2025mmg,xiong2025prune2drive,liu2025video}.
At the projector, adaptive pooling (DeCo~\cite{yao2024deco}) and query-based projectors (BLIP-2's Q-Former~\cite{li2023blip2}) shorten the sequence before it enters the language model. Within the LLM, methods include training-based policies guided by cross-modal supervision (e.g., LVPruning~\cite{sun2025lvpruning}, Skip-Vision~\cite{zeng2025skipvision}) and training-free heuristics that use attention or feature similarity to prune or merge tokens dynamically during inference (e.g., FastV~\cite{chen2024image}, SparseVLM~\cite{zhang2024sparsevlm}, DART~\cite{wen2025dart}, EfficientVLA~\cite{yang2025efficientvla}). However, since a significant portion of computation already occurs in the ViT, where the initial visual tokens are processed, compression applied only within the LLM can provide only limited acceleration, as the vision encoder remains a bottleneck for overall efficiency.

\noindent \textbf{Visual Compression in Visual Encoders.} 
% \paragraph{Visual Compression in Visual Encoders.} 
In visual encoders, token count can be reduced structurally or dynamically. Structurally, multi-scale encoders and hierarchical backbones with patch merging (e.g., LLaMA-VID~\cite{li2024llama}, Swin Transformer~\cite{liu2021swin}) progressively lower spatial or temporal resolution across stages, thus reducing the number of tokens. Dynamically, tokens are compressed within transformer layers: Token Merging (ToMe)~\cite{bolya2023tome} merges redundancy, while pruning within the ViT (DynamicViT~\cite{rao2021dynamicvit}, EViT~\cite{liang2022evit}, SparseViT~\cite{meng2022sparsevit}) removes low-importance tokens and reduces encoder computation. Most methods target single ViTs rather than MLLMs and often require extra training, which is costly in the MLLM setting.

% \section{Preliminaries}
% \input{Contents/Preliminary}

% To reduce vision-side computation while preserving semantic completeness for downstream multimodal reasoning, IPCV prunes redundant tokens early in the ViT and later reconstructs them via Neighbor-Guided Reconstruction.
%(NGR, see Algorithm~\ref{alg:ngr}).
\section{Methodology}
\label{sec:methodology}

\begin{figure*}[ht]
    \centering
    \includegraphics[width=\linewidth]{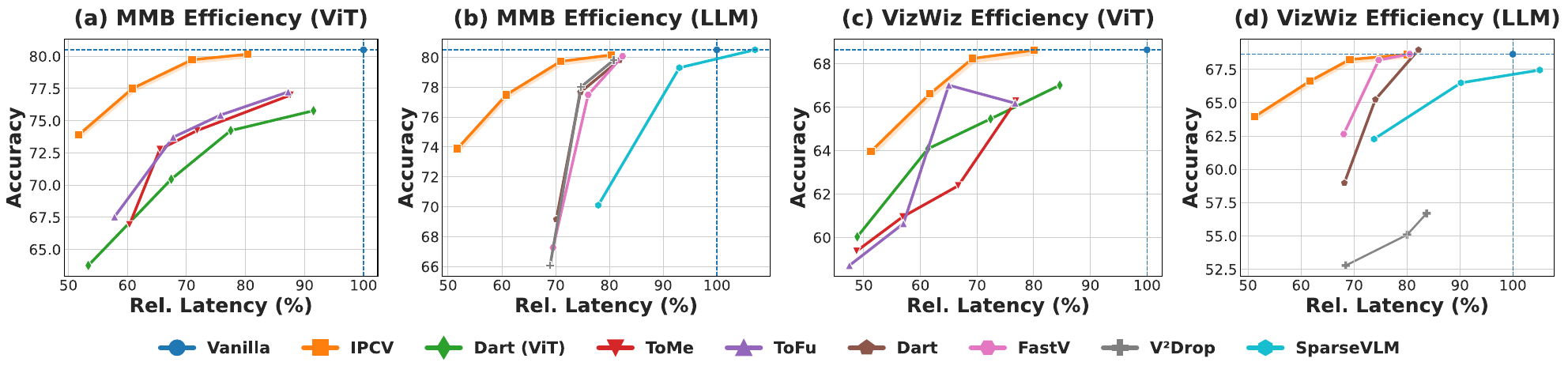}
    % \vspace{-2em}
    \caption{Efficiency comparison of IPCV vs. ViT/LLM baselines across datasets.
    (a) Accuracy vs. Rel.\ Latency for ViT baselines in MMB.
    (b) LLM baselines in MMB.
    (c) ViT baselines in VizWiz.
    (d) LLM baselines in VizWiz.
    \textbf{Rel.\ Latency} denotes total GPU inference time normalized to Vanilla. IPCV shows consistently strong efficiency--accuracy trade-offs on both datasets.}
    \label{fig:efficiency}
    \vspace{-1.25em}
\end{figure*}

Token compression in the vision encoder of MLLMs inherently risks discarding tokens that appear visually redundant but are semantically important. IPCV is designed to solve this problem (see Figure~\ref{fig:mylabel}): 
% In the early ViT layers, redundant tokens are pruned to reduce computation, while their semantic contribution is maintained through Neighbor-Guided Reconstruction, which rectifies the pruned tokens so they are still able to continue contributing in subsequent layers. 
In the early ViT layers, redundant tokens are pruned to reduce computation. Their semantic contribution is maintained through Neighbor-Guided Reconstruction, which rectifies the pruned tokens, allowing them to continue contributing in subsequent layers.
In this way, the LLM ultimately receives a complete and semantically intact set of vision tokens.

\subsection{Token Pruning in the Vision Encoder}

Let $l_p$ denote the index of the transformer block in the vision encoder 
where token pruning is applied.  
Given the input patch embeddings to block $l_p$, denoted as 
$\mathbf{H}_{l_p} \in \mathbb{R}^{L \times D}$, together with the embeddings 
from the preceding block $\mathbf{H}_{l_p-1}$, we compute a token-wise 
importance score $s_i$ based on the simple \emph{feature-difference} criterion:
\begin{equation}
    s_i = \left\| \mathbf{h}_{i,\,l_p} - \mathbf{h}_{i,\,(l_p-1)} \right\|_2, 
    \quad i = 1,\dots,L.
\end{equation}
Let $\mathcal{I}_{\text{keep}}$ be the indices of the top-$K$ tokens with the largest $s_i$, 
and $\mathcal{I}_{\text{rem}}$ be the remaining indices. 
The retained and removed token embeddings at the pruning layer $l_p$ are:
% \begin{equation}
%     \mathbf{H}_{\text{keep},\,l_p} = \mathbf{H}_{l_p}[\mathcal{I}_{\text{keep}}, :], \quad
%     \mathbf{H}_{\text{rem},\,l_p} = \mathbf{H}_{l_p}[\mathcal{I}_{\text{rem}}, :].
% \end{equation}
\begin{equation}
    \mathbf{H}_{\text{keep},\,l_p} = \mathbf{H}_{l_p}[\mathcal{I}_{\text{keep}}, :], \quad
    \mathbf{H}_{\text{rem},\,l_p} = \mathbf{H}_{l_p}[\mathcal{I}_{\text{rem}}, :].
\end{equation}

\subsection{Neighbor-Guided Reconstruction (NGR)}
We visualize the behavior of pruned tokens and their nearest neighbors in Figure~\ref{fig:trajectory_frequency}.
% Panel (a) shows that across subsequent layers, the hidden state trajectory of a pruned token remains highly consistent with the mean trajectory of its top-$10$ most similar unpruned tokens.
Panel (a) shows that, in the unpruned model, the hidden state trajectory of a token expected to be pruned remains highly consistent with the mean trajectory of its top‑10 most similar unpruned tokens.
This consistency indicates that the evolution of pruned tokens can be well approximated by the updates of their similar neighbors. 
Panels (b) and (c) present the empirical distribution of L1 distances between pruned and kept tokens, demonstrating that the change of a pruned token is much closer to the mean change of its top-$10$ similar tokens than to that of randomly selected tokens. 
These findings support the design of our Neighbor-Guided Reconstruction mechanism.

NGR aims to reconstruct the removed tokens at a layer $l > l_p$ by transferring local updates from their nearest retained neighbors, thereby enabling pruned tokens to continue contributing contextual information either within current attention layer or subsequent processing stages.

Given the retained and removed index sets $\mathcal{I}_{\text{keep}}$ and $\mathcal{I}_{\text{rem}}$ 
(and their embeddings $\mathbf{H}_{\text{keep},\,l_p}$ and $\mathbf{H}_{\text{rem},\,l_p}$), 
we can construct the rectified embeddings $\tilde{\mathbf{H}}_{\text{rem},\,l}$ 
for any given layer $l > l_p$ in the vision transformer.

% \paragraph{Neighbor selection.}
\noindent \textbf{Neighbor selection.}
For each removed token $i \in \mathcal{I}_{\text{rem}}$ with feature $\mathbf{h}_{i,\,l_p}$, 
we find its $k$ nearest neighbors among $\mathbf{H}_{\text{keep},\,l_p}$ 
% using the Euclidean distance:
by minimizing the Euclidean distance:
\begin{equation}
    \mathcal{N}_k(i) = \arg\min_{\substack{\mathcal{S} \subset \mathcal{I}_{\text{keep}}\\ |\mathcal{S}|=k}} 
    \sum_{j \in \mathcal{S}} \left\|\mathbf{h}_{i,\,l_p} - \mathbf{h}_{j,\,l_p}\right\|_2.
\end{equation}

% \paragraph{Local update transfer.}
\noindent \textbf{Local update transfer.}
% Let $\mathbf{H}_{\text{keep},\,l}$ be the retained-token embeddings at the current layer $l$, and define the per-token update on the kept set as:
Let $\mathbf{H}_{\text{keep},\,l}$ be the retained-token embeddings at the current layer $l$, and define the per-token update for the kept set from $l_p$ to $l$ as:
\begin{equation}
    \Delta \mathbf{H}_{\text{keep},\,l} = \mathbf{H}_{\text{keep},\,l} - \mathbf{H}_{\text{keep},\,l_p}.
\end{equation}
% The reconstruction of the removed token $i$ is then:
We reconstruct each removed token $i$ by adding the average updates of its $k$ neighbors to its original embedding $\mathbf{h}_{i,\,l_p}$:
\begin{equation}    
    \tilde{\mathbf{h}}_{i,\,l} = \mathbf{h}_{i,\,l_p} + \frac{1}{k} \sum_{j \in \mathcal{N}_k(i)} \Delta \mathbf{h}_{j,\,l}.
\end{equation}
Stacking all removed tokens yields $\tilde{\mathbf{H}}_{\text{rem},\,l} \in \mathbb{R}^{|\mathcal{I}_{\text{rem}}| \times D}$.
The complete NGR procedure is presented in Algorithm~\ref{alg:ngr}.

% \vspace{-0.8em}
\begin{algorithm}[h]
\caption{Neighbor-Guided Reconstruction (NGR)}
\label{alg:ngr}
\begin{algorithmic}[1]
\STATE \textbf{Input:} $\mathcal{I}_{\text{keep}}, \mathcal{I}_{\text{rem}}, \mathbf{H}_{\text{keep},\,l_p}, \mathbf{H}_{\text{rem},\,l_p}, \mathbf{H}_{\text{keep},\,l}, k$
\STATE \textbf{Output:} $\tilde{\mathbf{H}}_{\text{rem},\,l}$
\FOR{each $i \in \mathcal{I}_{\text{rem}}$}
    \STATE Find $\mathcal{N}_k(i) \subset \mathcal{I}_{\text{keep}}$ of size $k$ minimizing $\sum_{j \in \mathcal{N}_k(i)} \|\mathbf{h}_{i,\,l_p} - \mathbf{h}_{j,\,l_p}\|_2$
\ENDFOR
\STATE $\Delta \mathbf{H}_{\text{keep},\,l} \gets \mathbf{H}_{\text{keep},\,l} - \mathbf{H}_{\text{keep},\,l_p}$
\FOR{each $i \in \mathcal{I}_{\text{rem}}$}
    \STATE $\tilde{\mathbf{h}}_{i,\,l} \gets \mathbf{h}_{i,\,l_p} + \frac{1}{k} \sum_{j \in \mathcal{N}_k(i)} \Delta \mathbf{h}_{j,\,l}$
\ENDFOR
\STATE Stack all $\tilde{\mathbf{h}}_{i,\,l}$ to form $\tilde{\mathbf{H}}_{\text{rem},\,l}$
\end{algorithmic}
\end{algorithm}

\begin{table*}[!t]
  \centering
  \caption{Comparison of methods on Qwen2-VL. 
  \textbf{Avg. Acc.} denotes the average percentage of performance relative to Vanilla; 
  \textbf{Rel.\ Latency} shows total GPU inference time normalized to Vanilla (100\%), measured on MMBench-EN.}
  \resizebox{0.98\textwidth}{!}{
  \begin{tabular}{lccccccccccc}
  \toprule
  \textbf{Methods} 
    & \textbf{GQA} 
    & \textbf{MMB} 
    & \textbf{MMB-CN} 
    & \textbf{MME} 
    & \textbf{POPE} 
    & \textbf{SEED} 
    & \textbf{VQA\textsuperscript{text}} 
    & \textbf{VizWiz} 
    & \textbf{OCRBench} 
    & \textbf{Avg. Acc.}$\uparrow$ 
    & \textbf{Rel.\ Latency}$\downarrow$  \\
  \midrule
  Vanilla 
    & 61.5 & 80.5 & 78.7 & 2319 & 89.0 & 76.6 & 82.1 & 68.6 & 80.3 
    & 100\% & \textcolor{level4}{100\%} \\
  
  \midrule
  \rowcolor{gray!20}
  \multicolumn{12}{c}{\textit{ViT retain 50\% tokens / LLM retain 50\% tokens}} \\
  \addlinespace
  
  DART (ViT)   
    & 57.6 & 74.2 & 72.7 & 2111 & 84.3 & 68.9 & 63.8 & 65.5 & 47.1 
    & 87.3\% & \textcolor{level3}{77.5\%} \\
  ToMe          
    & 59.1 & 74.2 & 73.5 & 2009 & 88.0 & 73.1 & 61.2 & 62.4 & 39.3 
    & 86.3\% & \textcolor{level3}{71.8\%} \\
  ToFu  
    & 59.0 & 75.4 & 73.9 & 1931 & \textbf{88.3} & 73.4 & 60.5 & 67.0 & \textbf{76.9} 
    & 92.1\% & \textcolor{level3}{75.8\%} \\

    \textbf{IPCV}          
    & \textbf{60.5} & \textbf{79.7} & \textbf{76.7} & \textbf{2253} & \textbf{88.3} & \textbf{74.4} & \textbf{79.6} & \textbf{68.2} & 76.3 
    & \textbf{97.8\%} & \textcolor{level3}{70.9\%} \\
  
  \midrule
  \rowcolor{gray!20}
  \multicolumn{12}{c}{\textit{ViT retain 35\% tokens / LLM retain 35\% tokens}} \\
  \addlinespace
  
  DART (ViT)   
    & 55.1 & 70.5 & 69.2 & 1966 & 81.3 & 64.1 & 59.1 & 64.1 & 41.2 
    & 82.4\% & \textcolor{level2}{67.4\%} \\
  ToMe          
    & 58.2 & 72.8 & 71.6 & 1844 & 87.5 & 71.6 & 55.4 & 61.0 & 30.2 
    & 82.4\% & \textcolor{level2}{65.5\%} \\
  ToFu  
    & 58.7 & 73.7 & 72.7 & 1838 & \textbf{87.8} & 71.6 & 56.3 & 60.6 & 31.2 
    & 82.9\% & \textcolor{level2}{67.8\%} \\

    \textbf{IPCV}          
    & \textbf{58.8} & \textbf{77.5} & \textbf{75.3} & \textbf{2203} & 87.5 & \textbf{72.7} & \textbf{76.5} & \textbf{66.6} & \textbf{70.8} 
    & \textbf{94.9\%} & \textcolor{level2}{60.8\%} \\
  
  \midrule
  \rowcolor{gray!20}
  \multicolumn{12}{c}{\textit{ViT retain 20\% tokens / LLM retain 20\% tokens}} \\
  \addlinespace
  
  DART (ViT)   
    & 50.0 & 63.8 & 63.0 & 1706 & 75.4 & 56.2 & 52.9 & 60.0 & \textbf{29.1} 
    & 73.4\% & \textcolor{level1}{53.4\%} \\
  ToMe          
    & 56.2 & 66.9 & 67.0 & 1733 & 85.3 & \textbf{68.0} & 49.2 & 59.4 & 18.5 
    & 76.5\% & \textcolor{level2}{60.4\%} \\
  ToFu  
    & \textbf{56.3} & 67.5 & 67.7 & 1703 & \textbf{85.9} & \textbf{68.0} & 49.5 & 58.7 & 16.7 
    & 76.3\% & \textcolor{level1}{57.8\%} \\
\textbf{IPCV}          
    & 55.8 & \textbf{73.9} & \textbf{73.5} & \textbf{1818} & 85.3 & 65.5 & \textbf{61.2} & \textbf{64.0} & 23.7 
    & \textbf{81.4\%} & \textcolor{level1}{51.7\%} \\
  
  \midrule
  \rowcolor{gray!20}
  \multicolumn{12}{c}{\textit{ViT retain 100\% tokens / LLM retain 20\% tokens}} \\
  \addlinespace
  DART          
    & 55.3 & 77.7 & 69.7 & 2035 & 83.9 & 67.5 & 66.0 & 65.2 & 46.0 
    & 86.4\% & \textcolor{level3}{74.6\%} \\
  FastV         
    & \textbf{58.6} & 77.5 & 76.5 & 2266 & 86.8 & 70.9 & \textbf{78.8} & \textbf{68.2} & \textbf{57.5} 
    & \textbf{93.7\%} & \textcolor{level3}{76.1\%} \\
  V$^2$Drop 
    & 55.1 & 78.0 & 76.2 & 2265 & \textbf{88.3} & \textbf{71.8} & 56.2 & 55.1 & 13.3 
    & 82.1\% & \textcolor{level3}{74.7\%} \\
  SparseVLM    
    & 58.1 & \textbf{79.3} & \textbf{76.9} & \textbf{2277} & 86.7 & 71.4 & 73.6 & 66.5 & 50.3 
    & 92.1\% & \textcolor{level4}{93.0\%} \\
  
  \midrule
  \rowcolor{gray!20}
  \multicolumn{12}{c}{\textit{ViT retain 100\% tokens / LLM retain 5\% tokens}} \\
  \addlinespace
  DART          
    & 49.1 & 69.2 & 54.2 & 1764 & 77.7 & 55.6 & 40.4 & 59.0 & 33.6 
    & 71.9\% & \textcolor{level3}{70.1\%} \\
  FastV         
    & \textbf{50.2} & 67.3 & 67.9 & 1872 & 76.1 & \textbf{60.8} & \textbf{57.9} & \textbf{62.6} & \textbf{34.4} 
    & \textbf{78.0\%} & \textcolor{level2}{69.5\%} \\
  V$^2$Drop 
    & 47.9 & 66.1 & 67.4 & \textbf{1989} & \textbf{82.2} & 59.4 & 51.5 & 52.8 & 16.9 
    & 73.5\% & \textcolor{level2}{69.0\%} \\
  SparseVLM    
    & 49.3 & \textbf{70.1} & \textbf{69.0} & 1876 & 75.0 & 60.4 & 43.7 & 62.3 & 21.0 
    & 74.3\% & \textcolor{level3}{77.9\%} \\
  
  \bottomrule
  \end{tabular}
  }
  \label{tab:image_benchmarks}
\end{table*}

% \subsection{Mitigating Attention Disruption}
\subsection{Attention Stabilization (AS)}

% Pruning at $l_p$ inevitably alters the attention distribution, as removed tokens no longer participate in query-key interactions. 
% To mitigate this with minimal computation, IPCV keeps pruned tokens as keys and values for several subsequent layers, letting them contribute context in attention while skipping the costly FFN layers~\cite{zeng2025skipvision}.
Pruning at $l_p$ alters the attention distribution because the removed tokens no longer participate in query-key interactions. 
To mitigate this with minimal overhead, IPCV temporarily retains the pruned tokens as keys and values for a few layers after pruning. Thus, they contribute context in attention while skipping the costly FFN layers~\cite{zeng2025skipvision}.

Concretely, for layers 
$l \in [\,l_p,\, l_p + \Delta l_{\max})$, 
the removed tokens are temporarily rectified via:
\begin{equation}
    \tilde{\mathbf{H}}_{\text{rem},\,l} = \mathrm{NGR}\big(\mathbf{H}_{\text{rem},\,l_p}, \mathbf{H}_{\text{keep},\,l_p}, \mathbf{H}_{\text{keep},\,l}\big).
\end{equation}
% where $\mathbf{H}_{\text{keep},\,l}$ are the retained-token embeddings at the current layer $l$. 
At $l = l_p$, this is equivalent to directly using $\mathbf{H}_{\text{rem},\,l_p}$.

Then, we restore the full token sequence $\mathbf{H}_{\text{full},\,l} \in \mathbb{R}^{L \times D}$ by placing both retained and reconstructed tokens back to their corresponding original positions:
% \begin{equation}
%     \mathbf{H}_{\text{full},\,l}[\mathcal{I}_{\text{keep}}, :] = \mathbf{H}_{\text{keep},\,l}, \quad
%     \mathbf{H}_{\text{full},\,l}[\mathcal{I}_{\text{rem}}, :] = \tilde{\mathbf{H}}_{\text{rem},\,l}.
% \end{equation}
\begin{equation}
\begin{aligned}
    \mathbf{H}_{\text{full},\,l}[\mathcal{I}_{\text{keep}}, :] 
        &= \mathbf{H}_{\text{keep},\,l}, \\
    \mathbf{H}_{\text{full},\,l}[\mathcal{I}_{\text{rem}}, :] 
        &= \tilde{\mathbf{H}}_{\text{rem},\,l}.
\end{aligned}
\end{equation}
The full sequence is provided as input to the multi-head attention module for bidirectional token interactions:
\begin{equation}
    \mathbf{H}'_{\text{full},\,l} = \mathrm{Attention}\big(\mathbf{H}_{\text{full},\,l}, \mathbf{H}_{\text{full},\,l}, \mathbf{H}_{\text{full},\,l}\big),
\end{equation}
after which only the updated retained tokens are kept and forwarded to the FFN:
\begin{equation}
    \mathbf{H}'_{\text{keep},\,l} = \mathbf{H}'_{\text{full},\,l}[\mathcal{I}_{\text{keep}}, :].
\end{equation}

This strategy is motivated by Skip-Vision~\cite{zeng2025skipvision}, which shows that in transformer-based MLLMs, the feed-forward network (FFN) dominates computation for visual tokens, while attention is relatively lightweight. By enabling pruned tokens to skip the FFN but still join attention, IPCV preserves semantic aggregation with negligible extra FLOPs.

\subsection{Reintegration at the Final ViT Layer}

After the final ViT block $l_{\text{final}}$, IPCV merges the updated retained tokens 
$\mathbf{H}_{\text{keep},\,l_{\text{final}}+1} \in \mathbb{R}^{|\mathcal{I}_{\text{keep}}| \times D}$ (denoting the output of $l_{\text{final}}$) with the newly reconstructed removed tokens 
$\tilde{\mathbf{H}}_{\text{rem},\,l_{\text{final}}+1} \in \mathbb{R}^{|\mathcal{I}_{\text{rem}}| \times D}$ 
to restore the complete sequence:
\begin{equation}
\begin{aligned}
    \mathbf{H}_{\text{full},\,l_{\text{final}}+1}[\mathcal{I}_{\text{keep}}, :] 
        &= \mathbf{H}_{\text{keep},\,l_{\text{final}}+1}, \\
    \mathbf{H}_{\text{full},\,l_{\text{final}}+1}[\mathcal{I}_{\text{rem}}, :] 
        &= \tilde{\mathbf{H}}_{\text{rem},\,l_{\text{final}}+1}.
\end{aligned}
\end{equation}
The resulting $\mathbf{H}_{\text{full},\,l_{\text{final}}+1} \in \mathbb{R}^{L \times D}$ matches the original sequence length $L$ prior to pruning. 
This Reintegration step guarantees that the downstream LLM receives the same number and ordering of visual tokens as in the uncompressed setting, thereby preserving both interface compatibility and semantic integrity without introducing additional computational overhead in earlier layers.

\begin{table*}[!t]
  \centering
  \caption{Comparison of methods on video benchmarks with Qwen2-VL. \textbf{Rel.\ Latency} shows total GPU inference time normalized to Vanilla (100\%), measured on MVBench.}
  \resizebox{0.98\textwidth}{!}{
  \begin{tabular}{l ccc cccc cc}
  \toprule
  \multirow{2}{*}{\textbf{Methods}} 
  & \multirow{2}{*}{\textbf{MVBench}} 
  & \multirow{2}{*}{\textbf{EgoSchema}} 
  & \multirow{2}{*}{\textbf{MLVU}} 
  & \multicolumn{4}{c}{\textbf{VideoMME}} 
  & \multirow{2}{*}{\textbf{Avg. Acc.}$\uparrow$} 
  & \multirow{2}{*}{\textbf{Rel. Latency}$\downarrow$} \\ 
  \cmidrule(lr){5-8}
    &   &   &   & \textbf{Overall} & \textbf{Short} & \textbf{Medium} & \textbf{Long} &   &   \\
  \midrule
  Vanilla 
    & 66.1 & 62.0 & 59.8 & 57.7 & 70.4 & 54.6 & 48.0 
    & 100\% & \textcolor{level4}{100\%} \\
  
  \midrule
  \rowcolor{gray!20}
  \multicolumn{10}{c}{\textit{ViT retain 50\% tokens / LLM retain 50\% tokens}} \\
  \addlinespace
  
  DART (ViT) 
    & 57.8 & 55.2 & 53.6 & 49.7 & 57.2 & 48.2 & 43.8 
    & 87.6\% & \textcolor{level4}{87.0\%} \\
  ToMe 
    & 50.4 & 48.8 & 49.0 & 44.8 & 48.3 & 43.6 & 42.4 
    & 78.8\% & \textcolor{level3}{70.5\%} \\
  ToFu 
    & 50.5 & 48.5 & 48.9 & 44.6 & 49.0 & 43.4 & 41.4 
    & 78.5\% & \textcolor{level3}{73.9\%} \\
\textbf{IPCV} 
    & \textbf{64.0} & \textbf{58.4} & \textbf{57.8} & \textbf{55.1} & \textbf{67.6} & \textbf{51.7} & \textbf{46.1} 
    & \textbf{95.7\%} & \textcolor{level2}{65.9\%} \\
  
  \midrule
  \rowcolor{gray!20}
  \multicolumn{10}{c}{\textit{ViT retain 100\% tokens / LLM retain 20\% tokens}} \\
  \addlinespace
  DART 
    & 58.9 & 59.2 & 55.1 & 53.0 & 64.1 & 49.4 & 45.4 
    & 92.1\% & \textcolor{level2}{67.9\%} \\
  FastV 
    & 50.9 & 54.7 & 53.4 & 49.4 & 58.2 & 45.7 & 44.4 
    & 85.6\% & \textcolor{level3}{73.3\%} \\
  V$^2$Drop 
    & \textbf{62.1} & 58.6 & 55.5 & 53.5 & 63.7 & \textbf{51.0} & 45.9 
    & 93.4\% & \textcolor{level2}{69.9\%} \\
  SparseVLM 
    & 60.9 & \textbf{61.8} & \textbf{56.5} & \textbf{54.0} & \textbf{65.7} & 49.8 & \textbf{46.6} 
    & \textbf{94.5\%} & \textcolor{level4}{135.9\%} \\
  
  \bottomrule
  \end{tabular}
  }
  \label{tab:video_benchmarks}
\end{table*}

\begin{table*}[!t]
    \centering
    \caption{Comparison of different methods on InternVL3-38B. \textbf{Rel.\ Latency} shows total GPU inference time normalized to Vanilla (100\%), measured on MMBench-EN.}
    \resizebox{0.98\textwidth}{!}{
    \begin{tabular}{lcccccccccc}
    \toprule
    \textbf{Methods} 
      & \textbf{GQA} 
      & \textbf{MMB} 
      & \textbf{MMB-CN} 
      & \textbf{MME} 
      & \textbf{POPE} 
      & \textbf{VQA\textsuperscript{text}} 
      & \textbf{VizWiz} 
      & \textbf{OCRBench} 
      & \textbf{Avg. Acc.}$\uparrow$ 
      & \textbf{Rel.\ Latency}$\downarrow$ \\
    \midrule
    Vanilla 
      & 60.8 & 89.0 & 88.9 & 2467 & {90.6} & {83.8} & {69.3} & {86.0} 
      & {100\%} & \textcolor{level4}{100\%} \\
    
    \midrule
    \rowcolor{gray!20}
    \multicolumn{11}{c}{\textit{ViT retain 40\% tokens / LLM retain 40\% tokens}} \\
    \addlinespace
    
    DART (ViT)   
      & 60.8 & \textbf{85.0} & \textbf{85.4} & 2198 & 88.7 & 59.7 & 67.2 & \textbf{50.0} 
      & 88.1\%  & \textcolor{level2}{63.9\%} \\
    ToMe          
      & 60.2 & 82.2 & 80.9 & 2092 & 89.6 & 52.2 & 67.4 & 31.3 
      & 82.7\%  & \textcolor{level2}{61.7\%} \\
    ToFu  
      & 59.8 & 83.9 & 83.5 & 2074 & \textbf{90.1} & 52.7 & 67.7 & 31.5 
      & 83.4\%  & \textcolor{level2}{61.9\%} \\
    \textbf{IPCV+FastV}   
      & \textbf{61.2} & 84.5 & 83.2 & \textbf{2364} & 88.3 & \textbf{70.3} & \textbf{67.9} & 47.6 
      & \textbf{89.9\%}  & \textcolor{level2}{64.4\%} \\
    
    \midrule
    \rowcolor{gray!20}
    \multicolumn{11}{c}{\textit{ViT retain 100\% tokens / LLM retain 5\% tokens}} \\
    \addlinespace
    DART          
      & 51.7 & 71.1 & 70.5 & 1916 & 77.6 & 46.4 & 60.1 & \textbf{33.5} 
      & 73.6\%  & \textcolor{level3}{71.5\%} \\
    FastV         
      & \textbf{54.0} & \textbf{76.6} & \textbf{77.0} & \textbf{1977} & \textbf{82.7} & \textbf{55.1} & \textbf{65.9} & 30.1 
      & \textbf{78.6\%}  & \textcolor{level3}{70.7\%} \\
    V\textsuperscript{2}Drop 
      & 43.9 & 59.5 & 59.1 & 1697 & 67.3 & 17.1 & 53.4 &  9.7 
      & 57.1\%  & \textcolor{level3}{70.5\%} \\
    SparseVLM    
      & 51.4 & 74.6 & 73.5 & 1916 & 79.3 & 52.2 & 62.5 & 27.5 
      & 75.1\%  & \textcolor{level3}{73.4\%} \\
    
    \bottomrule
    \end{tabular}
    }
    \label{tab:internvl}
\end{table*}

% \subsection{Compatibility with LLM-side Pruning Methods}
\subsection{Compatibility with LLM-side Pruning}
\label{sec:llm-pruning}

A key property of IPCV is the strict preservation of the visual token sequence: after vision-side pruning and NGR-based reconstruction, the sequence of visual tokens passed to the LLM exactly matches the original uncompressed sequence in both length and ordering. This consistency in sequence structure ensures \emph{interface compatibility} with the multimodal fusion module as well as the LLM input layer, allowing IPCV to be seamlessly integrated into existing architectures without requiring any modifications.

% A key property of IPCV is that, after vision-side pruning and NGR-based reconstruction, the sequence of visual tokens passed to the LLM exactly matches the original uncompressed sequence in both length and ordering. This strict preservation of sequence structure ensures \emph{interface compatibility} with the multimodal fusion module as well as the LLM input layer, allowing IPCV to be seamlessly integrated into existing architectures without requiring any modifications. 

Consequently, IPCV can be seamlessly combined with existing LLM-side token pruning or acceleration techniques that assume a complete token set. For instance, methods such as FastV~\cite{chen2024image} and DART~\cite{wen2025dart} can be directly applied to the LLM input sequence to further reduce computation by pruning less informative tokens in the language model’s self-attention layers. 
% This modular design allows IPCV to function as a drop-in vision-side token compression module that complements orthogonal efficiency techniques on the language side, enabling joint optimization.
In practice, IPCV serves as a drop-in vision-side token compression module that complements language-side efficiency methods in an orthogonal manner.

\section{Experiments}
\subsection{Experimental Setup}

\noindent \textbf{Models and Baselines.}  
We evaluate IPCV primarily on Qwen2-VL-7B-Instruct~\cite{qwen2vl}, an instruction-tuned vision–language model, and additionally on InternVL3-38B~\cite{zhu2025internvl3} to verify generalization across architectures.
Baselines include training-free vision-side token compression methods (ToMe~\cite{bolya2023tome}, ToFu~\cite{kim2024token}) and LLM-side token pruning methods (FastV~\cite{chen2024image}, V$^2$Drop~\cite{chen2025variation}, SparseVLM~\cite{zhang2024sparsevlm}, DART~\cite{wen2025dart}). For a direct comparison, we also implement a vision-side variant of DART, denoted DART (ViT). 

\noindent \textbf{Evaluation Details.} 
To enable fair comparison between ViT-stage and LLM-stage methods, we report results using the overall acceleration ratio as the evaluation metric. 
For ViT-stage pruning methods, we adopt symmetric settings where both the ViT encoder and LLM decoder retain 50\%, 35\%, or 20\% of tokens. For LLM-stage pruning methods, we use asymmetric settings that retain 100\% of ViT tokens while pruning LLM tokens to 20\% or 5\%.
Benchmarks cover diverse image datasets (GQA~\cite{hudson2019gqa}, MMBench~\cite{liu2024mmbench}, MME~\cite{fu2023mme}, POPE~\cite{li2023evaluating}, SEED~\cite{li2024seedbench}, TextVQA~\cite{singh2019textvqa}, VizWiz~\cite{gurari2018vizwiz}, OCRBench\cite{liu2024ocrbench}) and video datasets (MVBench~\cite{li2024mvbench}, EgoSchema~\cite{mangalam2023egoschema}, MLVU~\cite{MLVU}, VideoMME~\cite{fu2025video}). 
% Please refer to Appendix~\ref{sec:appendix_evaluation} for more details.
Additional details on implementation and datasets are provided in Appendices~\ref{sec:appendix_evaluation} and ~\ref{sec:appendix_datasets}.
% Additional details on implementation and datasets are provided in the Appendices.

\subsection{Main Results}
Tables~\ref{tab:image_benchmarks} and~\ref{tab:video_benchmarks} summarize the accuracy--latency trade-off of IPCV and baselines on Qwen2-VL-7B across diverse benchmarks. 
Here, \textbf{Avg. Acc.} is the average percentage of performance relative to the vanilla model. 
\textbf{Rel. Latency} is total inference time normalized to vanilla (100\%), measured on MMBench-EN for image and MVBench for video. 
To highlight the speed--accuracy trade-off, latencies are color-coded by relative runtime: 
dark blue (\textcolor{level4}{$\geq80\%$}), 
blue-green (\textcolor{level3}{70--80\%}), 
green (\textcolor{level2}{60--70\%}), and
light green (\textcolor{level1}{50--60\%}).

On image understanding tasks, IPCV consistently achieves a superior balance between performance and efficiency. Under moderate acceleration (70.9\% vanilla runtime), it preserves 97.8\% of the vanilla performance, outperforming all baselines with comparable or higher runtime. 
% At a higher speedup of \textcolor{level2}{60.8\%} runtime, IPCV still retains 94.9\% accuracy, exceeding the next-best method by at least 12\% in Avg.\ Acc.
At a higher speedup of 60.8\% runtime, IPCV still retains 94.9\% Avg.\ Acc, exceeding the next-best method by at least 12 points.
% Even under the more aggressive \textcolor{level1}{51.7\%} runtime, it maintains 81.4\% accuracy.
In contrast, ViT-stage pruning baselines are prone to removing visual tokens that are critical for grounding textual understanding, resulting in pronounced accuracy degradation. 
% This issue is particularly evident on visual detail-sensitive tasks (OCRBench, VizWiz, TextVQA), where IPCV proves more stable than these vision-side methods.
This issue is particularly evident on visual detail-sensitive tasks (OCRBench, VizWiz, TextVQA), where IPCV proves more stable.
LLM-stage pruning methods generally maintain reasonable accuracy under modest acceleration, but their performance declines noticeably when aiming for higher speedups, indicating a limited ability to sustain accuracy at high compression ratios.

On video understanding tasks, IPCV attains 95.7\% of vanilla performance with lower inference latency, outperforming all baselines in this setting. Preserving text-critical visual cues also appears to be important for sustaining temporal-spatial reasoning. Similar to the image setting, ViT-stage pruning baselines suffer clear accuracy drops, while LLM-stage pruning methods, though relatively stable, still lag behind our method. IPCV reliably preserves multimodal reasoning across image and video benchmarks.

\subsection{Experiments on InternVL3}

To further examine the generalization of IPCV beyond Qwen2-VL, we conduct additional experiments on InternVL3-38B (Table~\ref{tab:internvl}). \textbf{Rel. Latency} is still measured on MMBench-EN for consistency. 
In this setting, we pair IPCV with FastV as the LLM-side pruner to test its overall effectiveness. The results show that IPCV continues to deliver favorable performance--efficiency trade-offs under this larger architecture: compared with alternative baselines, IPCV achieves higher accuracy at similar or lower runtime.

% \vspace{-0.3em}
\section{Analysis}
\label{sec:analysis}

\begin{table}[!ht]
    % \vspace{-1.0em}
    \centering
    \caption{Inference efficiency comparison on MMBench-EN. Latency columns show absolute time with relative percentage (normalized to Vanilla=100\%).}
    % \vspace{-0.5em}
    \resizebox{\linewidth}{!}{%\setlength{\tabcolsep}{4pt} % 缩小列间距
    \setlength{\tabcolsep}{1.5pt} % 缩小列间距
    % \resizebox{\textwidth}{!}{
    \begin{tabular}{lccccc}
    \toprule
    \multirow{2}{*}{\textbf{Methods}} & \textbf{Latency}$\downarrow$ & \textbf{Prefill Latency}$\downarrow$ & \textbf{FLOPs}$\downarrow$ & \textbf{KV Cache}$\downarrow$ & \textbf{Accuracy}$\uparrow$ \\
&(min:sec)&(min:sec)&(TFLOPs)&(MB)&(MMB)\\

    \midrule
    Vanilla & 61:12 {\footnotesize \textcolor{level4}{(100\%)}} & 51:18 {\footnotesize \textcolor{level4}{(100\%)}} & 31.9 & 76.6 & 80.5 \\
    \midrule
    \rowcolor{gray!20}
    \multicolumn{6}{c}{\textit{ViT retain 65\% tokens / LLM retain 65\% tokens}} \\
    \addlinespace
    
    DART (ViT) & 56:01 {\footnotesize \textcolor{level4}{(91.5\%)}} & 46:22 {\footnotesize \textcolor{level4}{(90.4\%)}} & 20.5 & 51.5 & 75.8 \\
    ToMe & 53:37 {\footnotesize \textcolor{level4}{(87.6\%)}} & 41:33 {\footnotesize \textcolor{level4}{(81.0\%)}} & 20.6 & 51.5 & 77.0 \\
    ToFu & 53:24 {\footnotesize \textcolor{level4}{(87.3\%)}} & 41:26 {\footnotesize \textcolor{level4}{(80.8\%)}} & 20.6 & 51.5 & 77.2 \\
    \textbf{IPCV} & 49:11 {\footnotesize \textcolor{level4}{(80.4\%)}} & 40:51 {\footnotesize \textcolor{level3}{(79.6\%)}} & 22.0 & 54.2 & 80.2 \\
    
    \midrule
    \rowcolor{gray!20}
    \multicolumn{6}{c}{\textit{ViT retain 50\% tokens / LLM retain 50\% tokens}} \\
    \addlinespace
    
    DART (ViT) & 47:26 {\footnotesize \textcolor{level3}{(77.5\%)}} & 37:24 {\footnotesize \textcolor{level3}{(72.9\%)}} & 16.0 & 40.8 & 74.2 \\
    ToMe & 43:57 {\footnotesize \textcolor{level3}{(71.8\%)}} & 34:20 {\footnotesize \textcolor{level2}{(66.9\%)}} & 16.2 & 40.7 & 74.2 \\
    ToFu & 46:22 {\footnotesize \textcolor{level3}{(75.8\%)}} & 34:30 {\footnotesize \textcolor{level2}{(67.3\%)}} & 16.2 & 40.7 & 75.4 \\
    \textbf{IPCV} & 43:25 {\footnotesize \textcolor{level3}{(70.9\%)}} & 33:46 {\footnotesize \textcolor{level2}{(65.8\%)}} & 18.0 & 44.6 & 79.7 \\
    
    \midrule
    \rowcolor{gray!20}
    \multicolumn{6}{c}{\textit{ViT retain 35\% tokens / LLM retain 35\% tokens}} \\
    \addlinespace
    
    DART (ViT) & 41:16 {\footnotesize \textcolor{level2}{(67.4\%)}} & 29:33 {\footnotesize \textcolor{level1}{(57.6\%)}} & 11.6 & 30.0 & 70.5 \\
    ToMe & 40:05 {\footnotesize \textcolor{level2}{(65.5\%)}} & 27:56 {\footnotesize \textcolor{level1}{(54.5\%)}} & 12.1 & 30.0 & 72.8 \\
    ToFu & 41:28 {\footnotesize \textcolor{level2}{(67.8\%)}} & 27:37 {\footnotesize \textcolor{level1}{(53.8\%)}} & 12.1 & 30.0 & 73.7 \\
    \textbf{IPCV} & 37:14 {\footnotesize \textcolor{level2}{(60.8\%)}} & 28:08 {\footnotesize \textcolor{level1}{(54.8\%)}} & 14.1 & 34.9 & 77.5 \\
    
    \midrule
    \rowcolor{gray!20}
    \multicolumn{6}{c}{\textit{ViT retain 20\% tokens / LLM retain 20\% tokens}} \\
    \addlinespace
    
    DART (ViT) & 32:40 {\footnotesize \textcolor{level1}{(53.4\%)}} & 27:28 {\footnotesize \textcolor{level1}{(53.5\%)}} & 7.5 & 19.3 & 63.8 \\
    ToMe & 36:56 {\footnotesize \textcolor{level2}{(60.4\%)}} & 27:28 {\footnotesize \textcolor{level1}{(53.5\%)}} & 8.3 & 19.3 & 66.9 \\
    ToFu & 35:23 {\footnotesize \textcolor{level1}{(57.8\%)}} & 27:20 {\footnotesize \textcolor{level1}{(53.3\%)}} & 8.3 & 19.3 & 67.5 \\
    \textbf{IPCV} & 31:39 {\footnotesize \textcolor{level1}{(51.7\%)}} & 26:42 {\footnotesize \textcolor{level1}{(52.1\%)}} & 10.5 & 25.3 & 73.9 \\
    
    \midrule
    \rowcolor{gray!20}
    \multicolumn{6}{c}{\textit{ViT retain 100\% tokens / LLM retain 35\% tokens}} \\
    \addlinespace
    DART & 50:07 {\footnotesize \textcolor{level4}{(81.9\%)}} & 40:34 {\footnotesize \textcolor{level3}{(79.1\%)}} & 20.4 & 34.9 & 79.8 \\
    FastV & 50:28 {\footnotesize \textcolor{level4}{(82.5\%)}} & 37:47 {\footnotesize \textcolor{level3}{(73.7\%)}} & 19.7 & 32.2 & 80.1 \\
    V$^2$Drop & 49:27 {\footnotesize \textcolor{level4}{(80.8\%)}} & 38:29 {\footnotesize \textcolor{level3}{(75.0\%)}} & 21.7 & 39.4 & 79.8 \\
    SparseVLM & 65:33 {\footnotesize \textcolor{level4}{(107.1\%)}} & 48:23 {\footnotesize \textcolor{level4}{(94.3\%)}} & 19.0 & 29.4 & 80.5 \\
    \midrule
    \rowcolor{gray!20}
    \multicolumn{6}{c}{\textit{ViT retain 100\% tokens / LLM retain 20\% tokens}} \\
    \addlinespace
    DART & 45:40 {\footnotesize \textcolor{level3}{(74.6\%)}} & 38:34 {\footnotesize \textcolor{level3}{(75.2\%)}} & 17.9 & 25.3 & 77.7 \\
    FastV & 46:33 {\footnotesize \textcolor{level3}{(76.1\%)}} & 33:45 {\footnotesize \textcolor{level2}{(65.8\%)}} & 17.1 & 22.6 & 77.5 \\
    V$^2$Drop & 45:43 {\footnotesize \textcolor{level3}{(74.7\%)}} & 34:40 {\footnotesize \textcolor{level2}{(67.6\%)}} & 18.6 & 27.9 & 78.0 \\
    SparseVLM & 56:56 {\footnotesize \textcolor{level4}{(93.0\%)}} & 41:08 {\footnotesize \textcolor{level4}{(80.2\%)}} & 16.5 & 20.3 & 79.3 \\
    \midrule
    \rowcolor{gray!20}
    \multicolumn{6}{c}{\textit{ViT retain 100\% tokens / LLM retain 5\% tokens}} \\
    \addlinespace
    DART & 42:55 {\footnotesize \textcolor{level3}{(70.1\%)}} & 30:10 {\footnotesize \textcolor{level1}{(58.8\%)}} & 15.4 & 15.9 & 69.2 \\
    FastV & 42:33 {\footnotesize \textcolor{level2}{(69.5\%)}} & 31:30 {\footnotesize \textcolor{level2}{(61.4\%)}} & 14.6 & 13.0 & 67.3 \\
    V$^2$Drop & 42:13 {\footnotesize \textcolor{level2}{(69.0\%)}} & 31:36 {\footnotesize \textcolor{level2}{(61.6\%)}} & 15.5 & 16.4 & 66.1 \\
    SparseVLM & 47:42 {\footnotesize \textcolor{level3}{(77.9\%)}} & 35:29 {\footnotesize \textcolor{level2}{(69.2\%)}} & 14.1 & 11.3 & 70.1 \\
    \bottomrule
    \end{tabular}
    % }
    }
    \label{tab:efficiency}
    \vspace{-1.5em}
\end{table}

% \begin{table*}[!t]
% \centering
% \caption{Combination of IPCV with different LLM-stage pruning methods. \textbf{Avg. Acc.} denotes the average percentage of performance relative to Vanilla.}
% % \vspace{-0.5em}
% % \setlength{\tabcolsep}{4pt} % 缩小列间距
% % \renewcommand{\arraystretch}{0.82} % 缩小行高
% \resizebox{0.98\textwidth}{!}{
% \begin{tabular}{lcccccccccc}
% \toprule
% \textbf{Methods} & \textbf{GQA} & \textbf{MMB} & \textbf{MMB-CN} & \textbf{MME} & \textbf{POPE} & \textbf{SEED} & \textbf{VQA\textsuperscript{text}} & \textbf{VizWiz} & \textbf{OCRBench} & \textbf{Avg. Acc.} \\
% \midrule
% Vanilla & 61.5 & 80.5 & 78.7 & 2319 & 89.0 & 76.6 & 82.1 & 68.6 & 80.3 & 100\% \\
% \midrule
% \rowcolor{gray!20}
% \multicolumn{11}{c}{\textit{ViT retain 35\% tokens / LLM retain 35\% tokens}} \\
% \addlinespace
% IPCV+DART (default) & 58.8 & 77.5 & 75.3 & 2203 & 87.5 & 72.7 & 76.5 & 66.6 & 70.8 & 94.9\% \\
% IPCV+FastV & 58.0 & 76.9 & 75.5 & 2205 & 87.0 & 71.5 & 74.1 & 66.9 & 64.0 & 93.3\% \\
% IPCV+V\textsuperscript{2}Drop & 50.1 & 70.3 & 69.1 & 2048 & 84.4 & 64.7 & 17.4 & 52.9 & 5.4 & 73.3\% \\
% IPCV+SparseVLM & 57.8 & 77.7 & 76.3 & 2220 & 87.0 & 71.9 & 74.0 & 66.2 & 62.6 & 93.3\% \\
% \bottomrule
% \end{tabular}
% }
% \label{tab:ipcv_llm_combo}
% % \vspace{-0.5em}
% \end{table*}

\begin{table*}[!t]
\centering
\caption{Combination of IPCV with different LLM-stage pruning methods. \textbf{Avg. Acc.} denotes the average percentage of performance relative to Vanilla.}
\resizebox{0.98\textwidth}{!}{
\begin{tabular}{lcccccccccc}
\toprule
\textbf{Methods} & \textbf{GQA} & \textbf{MMB} & \textbf{MMB-CN} & \textbf{MME} & \textbf{POPE} & \textbf{SEED} & \textbf{VQA\textsuperscript{text}} & \textbf{VizWiz} & \textbf{OCRBench} & \textbf{Avg. Acc.}$\uparrow$ \\
\midrule
Vanilla & 61.5 & 80.5 & 78.7 & 2319 & 89.0 & 76.6 & 82.1 & 68.6 & 80.3 & 100\% \\
\midrule
\rowcolor{gray!20}
\multicolumn{11}{c}{\textit{ViT retain 35\% tokens / LLM retain 35\% tokens}} \\
\addlinespace
IPCV+DART (default) & \textbf{58.8} & 77.5 & 75.3 & 2203 & \textbf{87.5} & \textbf{72.7} & \textbf{76.5} & 66.6 & \textbf{70.8} & \textbf{94.9\%} \\
IPCV+FastV & 58.0 & 76.9 & 75.5 & 2205 & 87.0 & 71.5 & 74.1 & \textbf{66.9} & 64.0 & 93.3\% \\
IPCV+V\textsuperscript{2}Drop & 50.1 & 70.3 & 69.1 & 2048 & 84.4 & 64.7 & 17.4 & 52.9 & 5.4 & 73.3\% \\
IPCV+SparseVLM & 57.8 & \textbf{77.7} & \textbf{76.3} & \textbf{2220} & 87.0 & 71.9 & 74.0 & 66.2 & 62.6 & 93.3\% \\
\bottomrule
\end{tabular}
}
\label{tab:ipcv_llm_combo}
\end{table*}

\begin{table*}[!t]
\centering
\caption{Ablation study of IPCV. 
\textbf{Rel. Latency} shows inference latency normalized to Vanilla (100\%), measured on the MMBench-EN. \textbf{Reint.} denotes Reintegration module.}
\resizebox{0.98\textwidth}{!}{
\begin{tabular}{lccccccccccc}
\toprule
\textbf{Methods} 
  & \textbf{GQA} 
  & \textbf{MMB} 
  & \textbf{MMB-CN} 
  & \textbf{MME} 
  & \textbf{POPE} 
  & \textbf{SEED} 
  & \textbf{VQA\textsuperscript{text}} 
  & \textbf{VizWiz} 
  & \textbf{OCRBench} 
  & \textbf{Avg. Acc.}$\uparrow$ 
  & \textbf{Rel. Latency}$\downarrow$ \\
\midrule
Vanilla 
  & 61.5 & 80.5 & 78.7 & 2319 & 89.0 & 76.6 & 82.1 & 68.6 & 80.3 
  & 100\% & \textcolor{level4}{100\%} \\

\midrule
\rowcolor{gray!20}
\multicolumn{12}{c}{\textit{ViT retain 35\% tokens / LLM retain 35\% tokens}} \\
\addlinespace
IPCV          
  & 58.8 & \textbf{77.5} & 75.3 & 2203 & 87.5 & \textbf{72.7} & \textbf{76.5} & \textbf{66.6} & 70.8 
  & \textbf{94.9\%} & \textcolor{level2}{60.8\%} \\
IPCV (w/o AS) 
  & \textbf{59.2} & 76.6 & 74.8 & 2210 & 86.9 & 72.3 & 74.4 & 66.4 & 69.0 
  & 94.1\% & \textcolor{level1}{56.1\%} \\
IPCV (w/o Reint.) 
  & 58.7 & 76.0 & \textbf{75.4} & \textbf{2211} & \textbf{87.6} & 72.5 & 75.4 & 66.3 & \textbf{72.4} 
  & 94.7\% & \textcolor{level2}{63.1\%} \\
IPCV (w/o AS, Reint.) 
  & 58.6 & 76.0 & 74.5 & 2197 & 87.4 & 71.9 & 73.7 & 65.7 & 69.0 
  & 93.6\% & \textcolor{level2}{60.2\%} \\
\bottomrule
\end{tabular}
}
\label{tab:ablation}
\end{table*}

\subsection{Inference Efficiency Analysis}
We evaluate efficiency on MMBench-EN, measuring total GPU inference latency and prefilling latency, together with FLOPs and KV cache usage (Table~\ref{tab:efficiency}). 
% IPCV achieves a favorable balance of accuracy and efficiency: as pruning ratios increase, total latency drops to 70.9\%, 60.8\%, and 51.7\% of the vanilla baseline, while still preserving strong accuracy relative to competing methods. 
IPCV achieves a favorable balance of accuracy and efficiency: latency drops to 70.9\%, 60.8\%, and 51.7\% of the vanilla baseline, while maintaining strong accuracy over competing methods.

As shown in Figure~\ref{fig:efficiency}, ViT-stage pruning achieves larger latency reductions but suffers sharper accuracy degradation, whereas LLM-stage methods plateaus around 69\% runtime and fails to deliver additional speedup under higher pruning ratios.
Prefilling latency decreases proportionally with total latency, and results on VizWiz confirm IPCV's effectiveness across benchmarks.
Compared with LLM-stage pruning, ViT-stage methods such as IPCV retain more LLM tokens and consequently incur larger KV caches, yet this overhead remains acceptable for practical deployment.
Moreover, FLOPs alone do not reliably predict latency, methods with similar FLOPs can yield different GPU times.

% \vspace{-0.3em}
\subsection{Compatibility in LLM-stage}
% \vspace{-0.3em}
Table~\ref{tab:ipcv_llm_combo} examines the compatibility of IPCV with representative LLM-stage token pruning methods. We observe that IPCV combined with FastV, DART and SparseVLM achieves comparable accuracy, all maintaining over 93\% of vanilla performance. 
IPCV+V$^2$Drop, on the other hand, shows more noticeable fluctuations across benchmarks. As seen in the InternVL3-38B results, V$^2$Drop's performance can be less stable under different architectures. While SparseVLM preserves accuracy, it often incurs higher inference latency, limiting practical efficiency. Thus, in practice we primarily consider FastV and DART as suitable LLM-stage token pruning methods to pair with IPCV.

\subsection{Ablation Study}
Table~\ref{tab:ablation} presents the ablation analysis of IPCV. 
Removing either the AS module or the Reintegration step slightly reduces accuracy, and omitting both causes a more noticeable drop. 
These mechanisms provide such benefits with only a marginal impact on runtime. The key reason is that pruned tokens are reused only in lightweight attention while skipping the more costly feed-forward layers. 
The results confirm that AS and Reintegration are indeed complementary, enhancing IPCV without sacrificing efficiency.

\begin{figure}[!h]
    % \vspace{-0.5em}
    \centering
    \includegraphics[width=\linewidth]{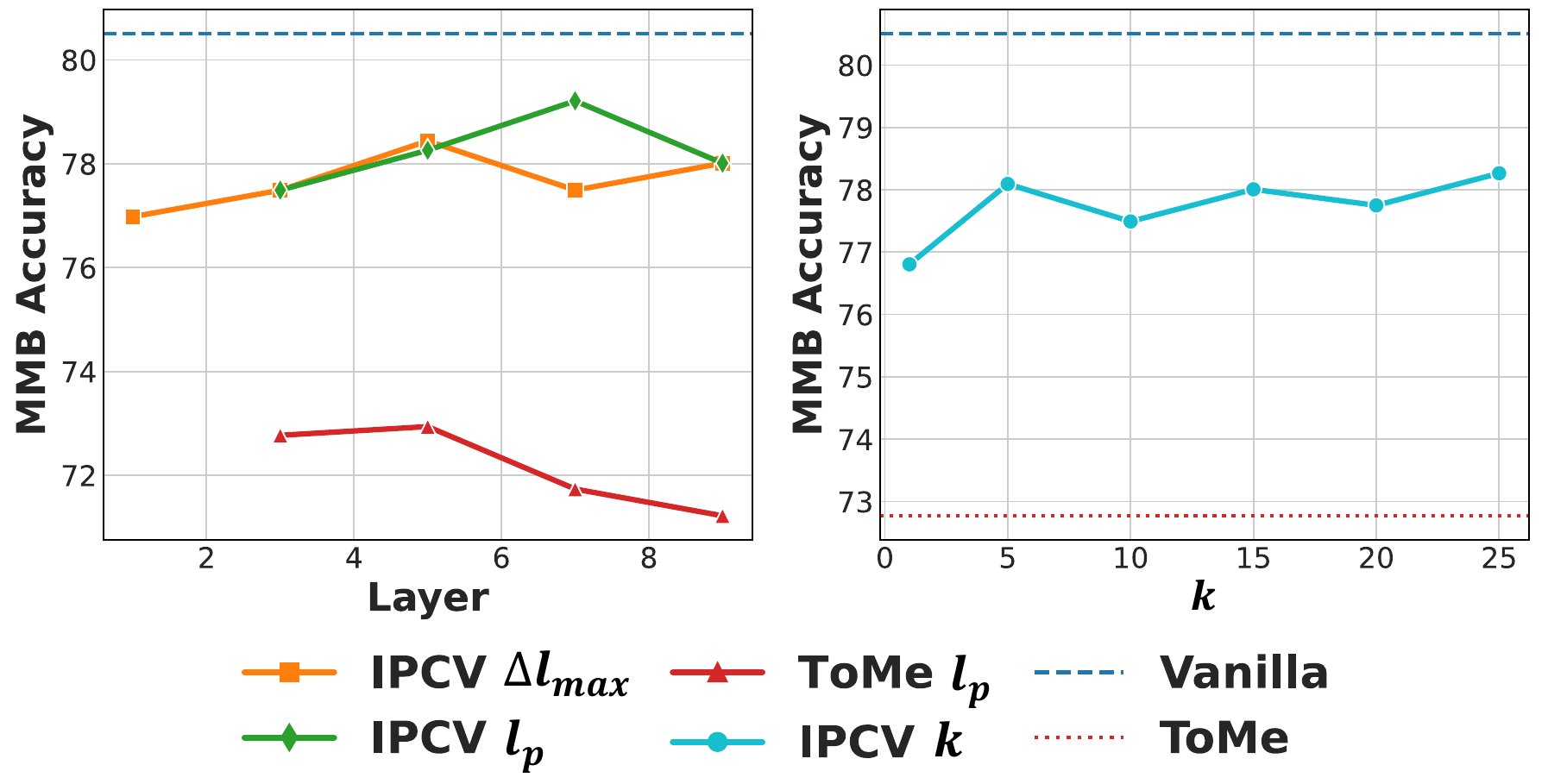}
    % \vspace{-1.5em}
    \caption{Sensitivity analysis of IPCV on MMBench-EN. 
    \textbf{(Left)} Impact of pruning start layer $l_p$ and AS depth $\Delta l_{\max}$. 
    \textbf{(Right)} Impact of neighbor size $k$ in NGR. 
    ToMe is included as a reference.}
    \label{fig:sensitivity}
    \vspace{-1.3em}
\end{figure}

\subsection{Sensitivity Analysis}
In our default configuration, the pruning start layer is $l_p=3$, the Attention Stabilization depth is $\Delta l_{\max}=7$, and the neighbor size in NGR is $k=10$. 
We evaluate IPCV's sensitivity to these hyperparameters on MMBench-EN (see Figure~\ref{fig:sensitivity}).
Varying the Attention Stabilization depth $\Delta l_{\max}$ shows that performance remains stable across a broad range (3--9).
This indicates that a moderate depth suffices for stabilizing attention, with IPCV delivering consistent performance across choices of this parameter.
For the pruning start layer $l_p$, later pruning is more accurate but less efficient, while earlier pruning is faster with a slight accuracy drop. 
We also analyze the sensitivity of the neighbor size $k$ in NGR. 
A very small $k$ can increase variance as reconstruction relies on too few neighbors, 
while a large $k$ may introduce noise from less similar tokens.
Empirically, IPCV remains stable over a wide range of $k$, confirming NGR’s robustness.
% Across all these settings, IPCV achieves better performance than other methods such as ToMe.

\section{Conclusion}
We introduce IPCV, a training-free framework for token compression in MLLM visual encoders. 
% By combining early-stage pruning with two NGR-based modules---Attention Stabilization and Reintegration---IPCV enables aggressive token reduction in the ViT while maintaining semantic completeness for downstream reasoning.
% IPCV combines early-stage pruning with two NGR-based modules: Attention Stabilization and Reintegration. This design enables aggressive token reduction in the ViT while maintaining semantic completeness for downstream reasoning.
IPCV combines early-stage pruning with two NGR-based modules: Attention Stabilization and Reintegration. This design enables aggressive token reduction in the ViT while preserving semantic information needed for downstream reasoning.
% Extensive experiments on diverse image and video benchmarks demonstrate that IPCV consistently achieves superior accuracy--efficiency trade-offs over existing training-free methods, while also generalizing well across different architectures. 
Experiments on diverse image and video benchmarks demonstrate that IPCV achieves superior accuracy--efficiency trade-offs compared to existing training-free methods. IPCV also generalizes well across different model architectures.
% Looking forward, IPCV provides a flexible foundation, and future work may focus on refining the reconstruction mechanism and deepening the analysis of Attention Stabilization, to further improve semantic preservation and computational efficiency, thereby enabling more powerful multimodal systems.
% Looking ahead, IPCV provides a flexible foundation. Future work could refine the reconstruction mechanism and further analyze the Attention Stabilization component, in order to improve semantic preservation and computational efficiency, enabling more powerful multimodal systems.
Looking ahead, IPCV provides a flexible foundation. Future work may focus on refining the reconstruction mechanism and further analyze the Attention Stabilization component to improve semantic preservation and computational efficiency, thereby enabling more powerful multimodal systems.

% \clearpage
%\input{Contents/limitation}

% References
{
    \small
    \bibliographystyle{ieeenat_fullname}
    \bibliography{main}
}

% \clearpage

% \input{author-kit-CVPR2026-v1-latex-/sec/X_suppl}

% \setcounter{page}{1}
% \maketitlesupplementary
\appendix

\section{Baseline Details}
\label{app_sec:baseline_details}

% To fairly evaluate our proposed IPCV framework, we compare it against representative baselines from two complementary perspectives. 
% \textbf{MLLM pruning methods} are originally designed for multimodal large language models (MLLMs), where they operate inside the language model by pruning less informative tokens from the multimodal input sequence.
% \textbf{ViT compression methods} were initially developed for vision-only transformers and tasks such as image classification. When applied to MLLMs, they operate directly on the vision encoder by merging or pruning visual tokens before projection into the multimodal space. 

To fairly evaluate our proposed IPCV framework, we compare it against representative baselines from two perspectives:
MLLM language-side pruning and ViT compression.

\subsection{MLLM Compression Methods}
% DART, V2Drop, FastV, SparseVLM

% These methods are originally designed for multimodal large language models (MLLMs), 
% where they operate inside the language model by pruning less informative tokens 
% from the multimodal input sequence. 

These methods are developed for MLLM token compression. 
They prune less informative tokens within the language model, without performing compression at the ViT stage.

% \paragraph{FastV.}
\noindent \textbf{FastV.}
FastV~\cite{chen2024image} prunes vision tokens in large vision-language models based on their attention scores. Tokens with low scores are removed early to accelerate inference.

\noindent \textbf{SparseVLM.}
SparseVLM~\cite{zhang2024sparsevlm} proposes a text-aware visual token sparsification framework for efficient vision-language model inference. It leverages text tokens as raters to assess the significance of visual tokens, pruning redundant ones with a recycling mechanism to minimize information loss.

\noindent \textbf{V$^2$Drop.}
V$^2$Drop~\cite{chen2025variation} introduces a variation-aware pruning strategy for large vision-language models. By measuring variation across layers, it adaptively discards uninformative tokens rather than relying on fixed importance scores.

\noindent \textbf{DART.}
DART~\cite{wen2025dart} leverages token similarity to identify and remove duplicated tokens. This approach does not rely on explicit attention scores, making it compatible with FlashAttention and avoiding large GPU memory overhead.

\subsection{ViT Compression Methods}
% ToMe, ToFu

ViT approaches are initially designed for vision-only transformers, targeting tasks like image classification. 
When applied to MLLMs, they operate directly on the vision encoder, merging or pruning visual tokens before the LLM-stage.

% When applied to MLLMs, they operate directly on the vision encoder, merging or pruning visual tokens prior to their projection into the multimodal space.

\noindent \textbf{ToMe.}
ToMe~\cite{bolya2023tome} accelerates vision transformers by merging similar tokens. Using a bipartite matching algorithm based on attention keys, it gradually combines redundant tokens across layers to shorten the sequence length.

\noindent \textbf{ToFu.}
ToFu~\cite{kim2024token} combines token pruning and merging in a unified framework. It adaptively chooses between them according to each layer’s characteristics and employs MLERP merging to better preserve feature norms.

\section{More Related Works}
\noindent \textbf{Multimodal Large Language Models.}
Multimodal large language models (MLLMs) extend LLMs beyond text to vision and other modalities, enabling VQA, captioning, and multimodal reasoning~\cite{liu2023llava, driess2023palme}. 
They typically couple a ViT pretrained with CLIP or SigLIP~\cite{radford2021clip, zhai2023sigmoid}, or Qwen2-VL's dynamic-resolution ViT~\cite{qwen2vl}, with a lightweight projector and an LLM. Higher-resolution images or longer temporal windows yield longer visual sequences: Qwen2-VL encodes a $308\times196$ image into 5{,}220 tokens in the ViT stage and 1{,}305 after merging~\cite{qwen2vl}; LLaVA maps $336\times336$ to 576 tokens, rising to 2{,}304 at $672\times672$~\cite{liu2023llava}; Video-LLaVA (8 frames) produces 2{,}048 tokens~\cite{lin-etal-2024-video}. 
The resulting abundance of tokens strains the vision encoder and cross-modal layers, increasing compute and latency.

\section{Implementation}
\label{sec:appendix_evaluation}

% \noindent \textbf{Evaluation Details.} 
% To enable fair comparison between ViT-stage and LLM-stage methods, we report results using the overall acceleration ratio as the evaluation metric. 
% For ViT-stage pruning methods, we adopt symmetric settings where both the ViT encoder and LLM decoder retain 50\%, 35\%, or 20\% of tokens. For LLM-stage pruning methods, we use asymmetric settings that retain 100\% of ViT tokens while pruning LLM tokens to 20\% or 5\%.
% Benchmarks cover diverse image datasets (GQA~\cite{hudson2019gqa}, MMBench~\cite{liu2024mmbench}, MME~\cite{fu2023mme}, POPE~\cite{li2023evaluating}, SEED~\cite{li2024seedbench}, TextVQA~\cite{singh2019textvqa}, VizWiz~\cite{gurari2018vizwiz}, OCRBench\cite{liu2024ocrbench}) and video datasets (MVBench, EgoSchema~\cite{mangalam2023egoschema}, MLVU~\cite{MLVU}, VideoMME~\cite{fu2025video}). Performance is reported relative to the uncompressed Vanilla model.

 % \noindent \textbf{Compression Settings.} 
 % We investigate various token retention ratios to assess performance under different compression levels. Our evaluation includes symmetric settings where both the ViT encoder and LLM decoder retain 50\%, 35\%, or 20\% of tokens, and asymmetric settings, such as retaining 100\% of ViT tokens while pruning LLM tokens to 20\% or 5\%. 
 % To ensure a fair comparison with multi-stage baselines, we normalize performance by the \textit{equivalent token count}---the average percentage of vision tokens preserved across all transformer layers.

% \noindent \textbf{Implementation Details.}
All experiments are conducted on Nvidia GPUs. 
Qwen2-VL-7B-Instruct is evaluated on image benchmarks using RTX 4090 (48GB), and on video benchmarks using A100-80G. 
InternVL3-38B is evaluated on image benchmarks using A100-80G. 
For ToMe and ToFu, we follow the original implementations with a minor modification: tokens are reduced proportionally across layers until the target sparsity is reached. 
Other baseline settings follow the original papers.

\section{Datasets}
\label{sec:appendix_datasets}
Our evaluation spans a diverse set of benchmarks for both image and video understanding.

\subsection{Image Datasets}

\noindent \textbf{GQA.}
GQA~\cite{hudson2019gqa} is built upon images, scene graphs, and compositional questions.
It provides detailed annotations of entities, attributes, and their relationships, together with diverse queries that require multi-step reasoning, spatial understanding, and logical inference.

\noindent \textbf{MMBench.}
MMBench~\cite{liu2024mmbench} comprises over 3,000 multiple-choice questions across 20 fine-grained ability dimensions such as object localization and social reasoning. Each dimension contains more than 125 questions.

\noindent \textbf{MME.}
MME~\cite{fu2023mme} contains 14 subtasks covering perception (e.g., object existence, count, OCR) and cognition (e.g., commonsense reasoning, calculation, translation). All instruction-answer pairs are manually constructed in concise yes/no format for straightforward evaluation.

\noindent \textbf{POPE.}
POPE~\cite{li2023evaluating} provides a polling-based evaluation benchmark designed to measure object hallucination in large vision-language models. It reformulates hallucination detection as a binary yes/no probing task about the presence of specific objects in an image.

\noindent \textbf{SEED.}
SEED~\cite{li2024seedbench} introduces a large-scale benchmark with human-annotated multiple-choice questions across 27 dimensions, providing hierarchical evaluation of MLLMs from image-text comprehension to joint text-image generation.

\noindent \textbf{TextVQA.}
TextVQA~\cite{singh2019textvqa} is built from natural images in Open Images containing textual elements, paired with human-posed questions requiring reading the embedded text. It is designed to test whether models can integrate OCR outputs with visual reasoning to answer text-centric questions.

\noindent \textbf{VizWiz.}
VizWiz~\cite{gurari2018vizwiz} is a goal-driven VQA dataset built from images taken and spoken questions asked by blind people using a mobile phone application. It is designed to evaluate models in realistic assistive scenarios, where images may be low quality, questions conversational, and some visual questions inherently unanswerable.

\noindent \textbf{OCRBench.}
OCRBench~\cite{liu2024ocrbench} evaluates large multimodal models on five OCR-related tasks, including text recognition, text-centric and document VQA, key information extraction, and handwritten expression recognition, aiming to expose their limitations on text-heavy vision tasks.

\subsection{Video Datasets}
\noindent \textbf{MVBench.}
MVBench~\cite{li2024mvbench} systematically converts static image tasks into dynamic, defining 20 temporal understanding tasks spanning perception to cognition. It provides multiple-choice QAs generated from 11 public video datasets.

\noindent \textbf{EgoSchema.}
EgoSchema~\cite{mangalam2023egoschema} is a benchmark of long egocentric video clips with multiple-choice questions, designed to test very long-form video-language understanding. It introduces temporal certificates to measure intrinsic temporal hardness and exposes limitations in long-term reasoning.

\noindent \textbf{MLVU.}
MLVU~\cite{MLVU} is built from diverse videos lasting minutes to hours across real and simulated domains. It defines nine tasks, such as summarization, action counting, and ordering, to evaluate models on complex long-video reasoning.

\noindent \textbf{Video-MME.}
Video-MME~\cite{fu2025video} contains 900 videos from six domains (e.g., knowledge, film \& television, and multilingual) with 2,700 multiple-choice questions. The videos range from 11 seconds to 1 hour with subtitles and audio.

\section{Theoretical Analysis}
This section provides a theoretical analysis of perturbations arising from ViT-stage pruning and Reintegration.

\begin{assumption}[Hausdorff-Lipschitz Continuity]\label{assump:hausdorff}
We assume the ViT mapping $F$ from layer $l_p$ to the output of $l_{\text{final}}$ is $L_{\mathrm{ViT}}$-Lipschitz with respect to the Hausdorff distance.
% Formally, for any two sets $\mathcal X, \mathcal Y \subset \mathbb{R}^d$, and any $i\in \mathcal X$ and $r\in \mathcal Y $,
% Formally, for any two sets $\mathcal X,\mathcal Y\subset\mathbb R^d$ and any indices $i\in \mathrm{Idx}(\mathcal X)$, $j\in\mathrm{Idx}(\mathcal Y)$, where $\mathrm{Idx}(\cdot)$ 
% denotes the index set of tokens and index correspondence is assumed to be preserved across layers, we have
Formally, for any two sets $\mathcal X,\mathcal Y\subset\mathbb R^d$ and any indices
$i\in \mathrm{Idx}(\mathcal X)$, $j\in\mathrm{Idx}(\mathcal Y)$, 
where $\mathrm{Idx}(\cdot)$ denotes the token index set, 
we assume index correspondence is preserved across layers. Then we have
\[
\|F_i(\mathcal X)- F_j(\mathcal Y)\|_2 \le L_{\mathrm{ViT}}\, d_H(\mathcal X,\mathcal Y),
\]
where $F_i(\cdot)$ denotes the final-layer output at position $i$. $d_H$ denotes the Hausdorff distance under the Euclidean norm:
% where $d_H$ is the set distance induced by the Euclidean norm:
\[
d_H(\mathcal X,\mathcal Y) := \max \bigl\{
\sup_{x\in \mathcal X}\inf_{y\in \mathcal Y}\|x-y\|_2, 
\sup_{y\in \mathcal Y}\inf_{x\in \mathcal X}\|x-y\|_2 \bigr\}
.
\]
% \(
% d_H(\mathcal X,\mathcal Y) := \max \\ \left\{
% \begin{aligned}
% &\sup_{x\in \mathcal X}\inf_{y\in \mathcal Y}\|x-y\|_2, 
% &\sup_{y\in \mathcal Y}\inf_{x\in \mathcal X}\|x-y\|_2
% \end{aligned}
% \right\}.
% \)

\end{assumption}

% \begin{assumption}[Bounded embeddings]\label{assump:bounded}
% There exists a constant $B>0$ s.t.\ all token embeddings in layer $l_p$ satisfy $\|x\|_2\le B$.
% \end{assumption}

\begin{assumption}[Bounded embeddings]\label{assump:bounded}
There exists a constant $B>0$ s.t.\ all token embeddings at layer $l_p$ satisfy
\[
\|\mathbf h_{i,l_p}\|_2 \le B,\qquad \forall i.
\]
\end{assumption}

% \begin{assumption}[Bounded embeddings]\label{assump:bounded}
% There exists a constant $B>0$ s.t.\ all token embeddings at layer $l_p$ satisfy
% \[
% \|x\|_2 \le B,\qquad \forall x \in \mathcal X.
% \]
% \end{assumption}

% ---------------- Lemma ----------------

\begin{lemma}[Pointwise deviation bound]
Consider the full token set $\mathcal X$ at layer $l_p$, with $\mathcal Y\subset \mathcal X$ denoting the subset retained after pruning.
For any $i\in \mathrm{Idx}(\mathcal X)$ and $r\in \mathrm{Idx}(\mathcal Y)$, we define the token update by:
\[
\Delta_i := F_i(\mathcal X)-\mathbf h_{i,l_p},\qquad
\Delta_r := F_r(\mathcal Y)-\mathbf h_{r,l_p},
\]
Then the following bound holds:
\[
\|\Delta_i-\Delta_r\|_2
\le
\|\mathbf h_{i,l_p}-\mathbf h_{r,l_p}\|_2
+ L_{\mathrm{ViT}}\,d_H(\mathcal X,\mathcal Y).
\]
\end{lemma}

\begin{proof}
By Assumption~\ref{assump:hausdorff} we have, for 
% the chosen indices $i\in \mathrm{Idx}(\mathcal X)$ and $r\in \mathrm{Idx}(\mathcal Y)$,
any output token $F_i(\mathcal X)\in F(\mathcal X)$ and $F_r(\mathcal Y)\in F(\mathcal Y)$
\[
\|F_i(\mathcal X)-F_r(\mathcal Y)\|_2 \le L_{\mathrm{ViT}}\,d_H(\mathcal X,\mathcal Y).
\]
Therefore the deviation of token updates is bounded by
\[
\begin{aligned}
\|\Delta_i-\Delta_r\|_2
&= \|F_i(\mathcal X)-F_r(\mathcal Y) -(\mathbf h_{i,l_p}-\mathbf h_{r,l_p})\|_2 \\
&\le \|F_i(\mathcal X)-F_r(\mathcal Y)\|_2 + \|\mathbf h_{i,l_p}-\mathbf h_{r,l_p}\|_2 \\
&\le \|\mathbf h_{i,l_p}-\mathbf h_{r,l_p}\|_2 + L_{\mathrm{ViT}}\,d_H(\mathcal X,\mathcal Y),
\end{aligned}
\]
which proves the stated pointwise bound.
\end{proof}

% \begin{proof}
% By Assumption~\ref{assump:hausdorff}, $d_H(F(X),F(Y))\le L_{\mathrm{ViT}}\,d_H(X,Y)$.
% Consequently, for any output token $F_i(X)\in F(X)$ and any $F_r(Y)\in F(Y)$, we have
% \[
% \|F_i(X)-F_r(Y)\|_2 \le L_{\mathrm{ViT}}\,d_H(X,Y) + \mathrm{diam}\big(F(Y)\big).
% \]
% Under Assumption~\ref{assump:bounded} the diameter of $F(Y)$ satisfies $\mathrm{diam}(F(Y))\le 2B$, hence
% \[
% \|F_i(X)-F_r(Y)\|_2 \le L_{\mathrm{ViT}}\,d_H(X,Y) + 2B.
% \]
% Therefore, the deviation of token update is bounded by
% \[
% \begin{aligned}
% \|\Delta_i-\Delta_r\|_2
% &= \|F_i(X)-F_r(Y) -(\mathbf h_{i,l_p}-\mathbf h_{r,l_p})\|_2 \\
% &\le \|F_i(X)-F_r(Y)\|_2 + \|\mathbf h_{i,l_p}-\mathbf h_{r,l_p}\|_2 \\
% &\le \|\mathbf h_{i,l_p}-\mathbf h_{r,l_p}\|_2 + L_{\mathrm{ViT}}\,d_H(X,Y) + 2B,
% \end{aligned}
% \]
% as claimed.
% \end{proof}

\begin{remark} \label{remark:pointwise-bound}

By Assumption~\ref{assump:bounded}, both $\|\mathbf h_{i,l_p}-\mathbf h_{r,l_p}\|_2$ and $d_H(\mathcal X,\mathcal Y)$ are at most $2B$. 
Substituting these bounds into the pointwise inequality gives
\[
\|\Delta_i-\Delta_r\|_2 \le 2B\,(L_{\mathrm{ViT}}+1).
\]

\end{remark}

\begin{theorem}[NGR Reconstruction Error]
% The NGR reconstruction satisfies, for each removed token \(i\),
%admits the bound
For each removed token \(i\), the NGR reconstruction satisfies the bound
\[
\|\widetilde{\mathbf h}_{i,l_{\text{final}}+1} - \mathbf h_{i,l_{\text{final}}+1}\|_2
\;\le\; 2B\,(L_{\mathrm{ViT}}+1).
\]
Here $\widetilde{\mathbf h}_{i,l_{\text{final}}+1}$ denotes the reconstructed hidden state
of pruned token $i$ via NGR, and $\mathbf h_{i,l_{\text{final}}+1}$ the true hidden state without pruning.
Consequently, the Hausdorff distance between the uncompressed final-layer set $\mathcal X'$ and
the reconstructed set $\widehat{\mathcal X}'$ is bounded by
\[
d_H(\mathcal X',\widehat{\mathcal X}') \le 2B\,(L_{\mathrm{ViT}}+1).
\]
\end{theorem}

\begin{proof}[Sketch]
Recall that
\[
\widetilde{\mathbf h}_{i,l_{\text{final}}+1}
- \mathbf h_{i,l_{\text{final}}+1}
= \frac{1}{k}\sum_{r\in\mathcal N_k(i)}(\Delta_r - \Delta_i),
\]
By the triangle inequality,
\[
\|\widetilde{\mathbf h}_{i,l_{\text{final}}+1} - \mathbf h_{i,l_{\text{final}}+1}\|_2
\le \frac{1}{k} \sum_{r\in\mathcal N_k(i)} \|\Delta_r - \Delta_i\|_2.
\]
From the Remark~\ref{remark:pointwise-bound} we obtain the worst-case bound
\(\|\Delta_r-\Delta_i\|_2 \le 2B\,(L_{\mathrm{ViT}}+1)\) for any \(r\). Hence each term
in the sum is bounded by \(2B(L_{\mathrm{ViT}}+1)\), so the average yields the desired inequality.
\end{proof}

\section{Visual Examples of Token Compression}
% Figure~\ref{fig:visual} presents diverse image examples of IPCV compression in the ViT stage.
% Without text guidance, our method tends to preserve 
% % structurally complex and visually detailed regions, making fine-grained image content less affected by pruning-induced degradation.
% visually rich regions, alleviating pruning impact on fine-grained content.

\begin{figure}[!h]
    
    \centering
    \includegraphics[width=\linewidth]{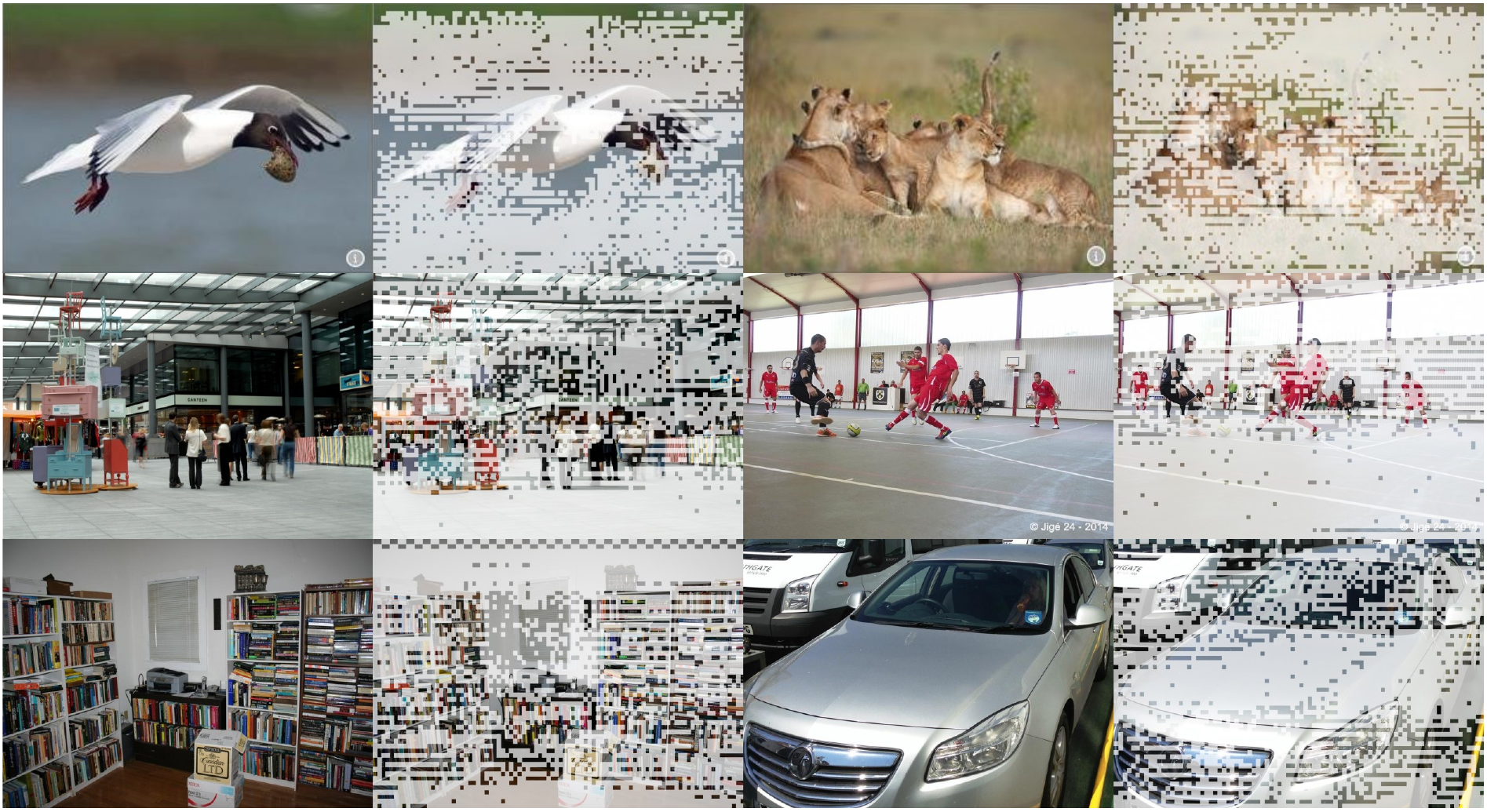}
    
    \caption{Original and compressed image pairs in the ViT-stage.}

    \label{fig:visual}
\end{figure}

Figure~\ref{fig:visual} presents diverse image examples of IPCV compression in the ViT stage.
Without text guidance, our method tends to remove less informative regions, alleviating the impact of pruning on fine-grained content.

% \clearpage

\end{document}